%% file: main.tex
\documentclass[10pt]{article} 

\usepackage[final]{ewrl_2026}
\usepackage[utf8]{inputenc} 
\usepackage[T1]{fontenc}    
\usepackage{hyperref}       
\usepackage{url}            
\usepackage{booktabs}       
\usepackage{amsfonts}       
\usepackage{nicefrac}       
\usepackage{microtype}      
\usepackage{xcolor}         
\usepackage{amsmath}
\usepackage{subcaption}     
\usepackage{graphicx}
\usepackage{enumitem}
\usepackage{makecell}

\title{NASimJax: A GPU-Accelerated Policy Learning Framework for Penetration Testing}

\author{
  Raphael Simon$^{1,2}$ \\
  \texttt{r.simon@cylab.be} \and
  José Carrasquel$^{1}$ \and
  Elli Makdis Antoun$^{1}$ \and
  Wim Mees$^{1}$ \and 
  Pieter Libin$^{2}$ \\ 
  \\
  $^{1}$CISS Department, Royal Military Academy, Belgium \\
  $^{2}$AI Lab, Vrije Universiteit Brussel, Belgium \\
}
\date{}

\begin{document}

\maketitle


\begin{abstract}
Penetration testing -- the practice of simulating cyberattacks to identify vulnerabilities -- is a complex sequential decision-making task that is inherently partially observable and features large action spaces. Existing RL simulators for this domain are CPU-bound and fixed to narrow scenarios, making it infeasible to train policies that generalize across networks. We present NASimJax, a JAX-native framework that formulates penetration testing as a Contextual POMDP and introduces a network generation pipeline producing structurally diverse, guaranteed-solvable scenarios. The framework reaches up to 80$\times$ higher environment throughput than previous simulators, enabling experiments on larger networks and tractable hyperparameter searches. We provide PPO and PQN baselines and conduct the first systematic evaluation of unsupervised environment design for penetration testing. We find that Prioritized Level Replay and ACCEL handle dense training distributions better than Domain Randomization, and that training on sparser topologies yields an implicit curriculum that improves generalization -- even to topologies denser than those seen during training. A recurrent PPO variant confirms that these distributional findings are not an artifact of feed-forward policies. The code is available at: \url{https://github.com/raphsimon/NASimJax}.
\end{abstract}


\section{Introduction}

Cybersecurity has become one of the most critical domains in modern society, underpinning healthcare, banking, and energy infrastructure. Among the tools used to proactively defend these systems is penetration testing -- the authorised practice of simulating attacks to identify vulnerabilities and misconfigurations in IT networks~\citep{nist800115}. While effective, penetration testing is laborious and expensive, requiring seasoned security experts to reason over long decision horizons, weigh competing attack vectors, and adapt to the specifics of each target network. 

Automating this process via reinforcement learning (RL) has only recently gained traction~\citep{liu2025autonomous}, motivated by the success of RL in other complex sequential decision-making domains~\citep{mnih2015human, openai2019dota, vinyals2019grandmaster}. The penetration testing problem maps naturally onto a Partially Observable Markov Decision Process (POMDP)~\citep{sarraute_pomdps_2012, sarraute_penetration_2013}. An agent cannot directly observe the full network topology or its vulnerabilities, and must actively acquire information before acting. 

Beyond partial observability, the domain compounds many of the difficulties common to real-world RL~\citep{dulac2021challenges} and specifically catalogued for pentesting by long decision horizons, sparse and delayed rewards, combinatorial action spaces, and strong dependence on network-specific structure~\citep{liu2025autonomous}. Together these properties make generalization the central deployment obstacle -- policies trained on narrow scenarios fail to transfer, and closing this gap requires simulators that expose agents to structural diversity and decision-relevant complexity rather than merely surface-level realism.

Existing penetration testing simulators~\citep{schwartz2019nasim, msft:cyberbattlesim, emerson2024cyborg++} fall short on two fronts. First, they are primarily fixed to single or narrowly parameterized scenarios, providing insufficient diversity to train generalizable policies. Second, they rely on CPU-bound Python implementations, which creates a fundamental throughput bottleneck: RL requires millions of environment interactions due to sample inefficiency, yet these simulators cannot generate experience fast enough to fully utilize modern accelerators. This bottleneck is especially costly in penetration testing, where RL algorithms are not yet well understood and thorough hyperparameter searches -- crucial for reliable results in any new domain, given the known sensitivity of RL to hyperparameter choices~\citep{patterson2024empirical, adkins2024method} -- are intractable at the speeds current tools offer. 

To address these challenges, we present NASimJax, a JAX-native research framework for studying generalization in automated penetration testing. NASimJax adopts NASim's framing of the task as partially observable sequential information gathering~\citep{schwartz2019nasim}, but redesigns the state representation, network generation, and context interface to support distributional training and unsupervised environment design (UED). Our contributions are:
\begin{itemize}[leftmargin=*, noitemsep, topsep=2pt]
  \item \textbf{Distributional formulation and generator.} We formulate penetration testing as a Contextual POMDP~\citep{hallak2015contextual, kirk_survey_2023}, where each episode is conditioned on a network instance sampled from a configurable distribution. This turns a historically fixed-scenario problem into a distributional one suited to studying zero-shot transfer. A parametric generation pipeline provides this distribution, producing structurally diverse, guaranteed-solvable networks with fine-grained control over topology and vulnerability density.
  \item \textbf{Tractable large-scale experimentation.} An end-to-end JIT-compiled and vectorized design reaches up to 80$\times$ the throughput of the most widely adopted existing simulator (Figure~\ref{fig:speed_comparison_memory_profile}). This makes previously impractical experiments feasible: large hyperparameter sweeps, billion-step training on 40-host networks, and regret-based UED through existing JAX libraries.
  \item \textbf{Baselines and the first UED study for penetration testing.} We provide PPO and PQN baselines on 16-, 26-, and 40-host networks. Using PPO and PPO-LSTM, we evaluate Domain Randomization (DR)~\citep{tobin_2017_DR}, Prioritized Level Replay (PLR)~\citep{jiang2021plr}, and ACCEL~\citep{parker-holder_2023_accel} for zero-shot transfer. Regret-based methods are more robust than DR under dense training distributions, and sparse-topology training produces an implicit curriculum that improves generalization, even to topologies denser than those seen during training.
\end{itemize}

\section{Contextual Reinforcement Learning}\label{sec:contextual_rl}

A Markov Decision Process (MDP)~\citep{puterman_1990_mdp} is a tuple $\langle \mathcal{S}, \mathcal{A}, \mathcal{P}, \mathcal{R}, \rho_0, \gamma \rangle$ with state space $\mathcal{S}$, action space $\mathcal{A}$, transition kernel $\mathcal{P}(s' \mid s,a)$, reward function $\mathcal{R}$, initial state distribution $\rho_0$, and discount factor $\gamma \in [0,1]$. The goal is to learn a policy $\pi(a \mid s)$ maximizing the expected return $\mathbb{E}_\pi[G_0]$, with $G_t = \sum_{k=0}^{\infty} \gamma^k r_{t+k}$. A Partially Observable MDP (POMDP)~\citep{sutton_rl_2018} extends this with an observation space $\Omega$ and observation function $\mathcal{O}(o \mid s',a)$: the agent perceives only $o_t$ and must act on the observation history, or equivalently on a belief state $b_t \in \Delta(\mathcal{S})$.

Following~\citet{hallak2015contextual} and~\citet{kirk_survey_2023}, a \emph{Contextual POMDP} augments this with a context $c \in \mathcal{C}$, sampled once per episode from $p(c)$, that parameterizes the dynamics, rewards, observations, and initial state: $\langle \mathcal{S}, \mathcal{A}, \mathcal{P}_c, \mathcal{R}_c, \Omega, \mathcal{O}_c, \rho_{0,c}, \gamma, \mathcal{C}, p(c) \rangle$. The context is fixed within an episode but varies across episodes, inducing a task distribution. The objective is $\mathbb{E}_{c \sim p(c)}\!\left[ \mathbb{E}_\pi\!\left[ G_0 \mid c \right] \right]$.

\section{Related Work}

\paragraph{JAX and Hardware-Accelerated RL.} 
JAX~\citep{jax2018github} enables JIT-compiled, functionally pure programs to run end-to-end on accelerators, and its \texttt{vmap} transformation allows environment \texttt{step} and \texttt{reset} functions to be vectorized across thousands of parallel instances. Together these eliminate the CPU--GPU bottleneck that limits conventional Python-based RL pipelines~\citep{rutherford2024jaxmarl}. A growing ecosystem of JAX-native RL environments exploits this, including Gymnax~\citep{gymnax2022github}, Craftax~\citep{matthews2024craftax}, and Jumanji~\citep{bonnet2024jumanji}.

\paragraph{Penetration Testing Simulators.}
Among RL-oriented simulators, NASim~\citep{schwartz2019nasim} is the most widely adopted and underpins much subsequent work, including the emulation-backed extensions NASimEmu~\citep{janisch_nasimemu_2023} and PenGym~\citep{nguyen2025pengym}, which bridge simulated training with virtualized targets. StochNASim~\citep{simon2025learningrobustpenetrationtestingpolicies} adds stochastic resets and variable network sizes on top of NASim. Beyond the NASim lineage, CyberBattleSim~\citep{msft:cyberbattlesim} models lateral movement, CybORG~\citep{standen_cyborg_2021} couples offensive and defensive agents within a shared network, and C-CyberBattleSim~\citep{terranova2025scalable} grounds scenarios in real-world vulnerability databases and scanner output. All are CPU-bound Python implementations, fundamentally limiting the scale at which distributional training can be conducted.

\paragraph{Generalization in Automated Penetration Testing.}
Generalization is a recognized open problem for RL-based penetration testing~\citep{liu2025autonomous}, and prior work has approached it along two axes. \emph{Architecturally}, permutation- and size-invariant networks have been proposed to make policies transferable across networks of varying size and host ordering~\citep{janisch_nasimemu_2023, terranova2025scalable}. \emph{Distributionally}, agents have been exposed to diverse training environments through multi-scenario training~\citep{janisch_nasimemu_2023}, stochastic resets with varying network sizes~\citep{simon2025learningrobustpenetrationtestingpolicies}, and Domain Randomization~\citep{zhou2024towards}; the recent survey of Liu et al.~\citep{liu2025autonomous} identifies DR as the only distributional-based technique applied in this domain to date. To our knowledge, no prior work has applied regret-based Unsupervised Environment Design methods such as PLR or ACCEL, leaving open whether explicit regret-based curricula can outperform DR in this setting. We provide the first empirical evidence on this question.

\section{NASimJax}\label{sec:nasimjax}

NASimJax preserves NASim's core task abstraction -- penetration testing as a sequential information-gathering problem under partial observability -- but the state representation, network generation pipeline, and context framework are redesigned from the ground up to support CMDP-style training, JIT compilation, and structural diversity. The environment is designed as a configurable research abstraction rather than a fixed benchmark scenario: its primary purpose is to support controlled variation over network instances so that policy robustness and zero-shot generalization can be studied systematically. Through a stochastic network generation process, the environment can be scaled in complexity and difficulty, allowing it to remain challenging as RL algorithms improve and preventing overfitting to narrowly defined scenarios. The environment adheres to the Gymnax API~\citep{gymnax2022github}, enabling integration with JAX-based RL algorithms and direct comparison of learning methods within a standardized environment.

\paragraph{Modeling scope.} NASimJax abstracts penetration testing at the level of attack-path discovery in segmented enterprise networks. The agent must chain reconnaissance, exploitation, and privilege escalation across hosts whose operating systems, services, and processes are initially hidden. Vulnerabilities are modeled as $\text{OS} \times \text{service}$ and $\text{OS} \times \text{process}$ pairs that admit specific exploits; hosts may also be correctly configured and run no exploitable software at all, capturing the uneven security posture of real deployments. Intentionally abstracted away are CVE-level semantics, exploit timing, detection and defender response, persistence mechanisms, and data exfiltration. This scope preserves the combinatorial and information-theoretic structure that makes penetration testing hard for RL (i.e., long horizons, large action spaces, partial observability, and strong dependence on network-specific structure) while remaining tractable enough for systematic study. The framework is suited to investigating attack-path discovery, action prioritization under uncertainty, and policy generalization across network topologies; it is not intended for questions about low-level exploit mechanics or active defender dynamics.

\paragraph{Task.} An episode proceeds as follows. An agent interacts with a network by scanning hosts, enumerating services, exploiting vulnerable software to obtain an initial foothold, and escalating privileges through vulnerable processes. Information about topology, running services, and vulnerabilities must be actively acquired through interaction. Progress is inherently sequential and long-horizon, as successful exploitation of certain hosts may be required to reach others. An episode terminates once the agent has obtained the required privileges on a predefined set of sensitive hosts. Each episode is conditioned on a learning context (scenario), represented by a concrete network instance comprising hosts partitioned into subnets, connectivity rules defining inter-subnet communication, and host-specific configurations (running services, processes, and associated exploitability). While the policy is shared across episodes, the underlying context varies, providing a principled mechanism to expose the agent to diverse network structures during training. This naturally supports viewing the environment as a Contextual POMDP, where generalization is achieved by learning policies that perform well across a distribution of network contexts rather than memorizing individual attack paths.

\subsection{Task specification}\label{sec:task-spec}
We briefly describe the state, action and observation spaces, and reward function. The core structure follows NASim~\citep{schwartz2019nasim} and subsequent work~\citep{janisch_nasimemu_2023, simon2025learningrobustpenetrationtestingpolicies}; the design contributions of NASimJax -- the state encoding, generation pipeline, context framework, and reward scaling -- are addressed in Sections~\ref{sec:network_generation}--\ref{sec:context_sets} and~\ref{sec:reward_scaling}. Full semantics are provided in Appendix~\ref{app:state_repr}.

\paragraph{State.} Network state comprises an adjacency matrix encoding inter-subnet traffic and firewall rules, together with per-host properties (subnet address, reachability, discovery status, access level, OS, vulnerable services, and vulnerable processes). Host properties are batched into a single $\{0,1\}^{N_h \times F}$ tensor using one-hot and boolean encodings (cf. Figure~\ref{fig:context_gen_batched_state}). This encoding reduces transition logic to bitwise operations and keeps every host property in a single byte, reducing the overall memory footprint. The full state is used internally for transition dynamics but is not exposed to the agent, preserving partial observability.

\paragraph{Actions.}\label{sec:action_space} The action space contains six discrete action types grouped into reconnaissance (OS, process, service, and subnet scans) and exploitation (exploit, privilege escalation). Each action targets a specific host, so the total number of environment actions is
\begin{equation}
|\mathcal{A}| = N_h \times \left(|\mathcal{A}_{\text{scan}}| + |\text{OS}| \times (|V_{\text{svc}}| + |V_{\text{proc}}|)\right),
\label{eq:action-space}
\end{equation}
where $N_h$ is the number of hosts, $|\mathcal{A}_{\text{scan}}|$ the number of scan actions, $|\text{OS}|$ the number of operating systems, and $|V_{\text{svc}}|, |V_{\text{proc}}|$ the numbers of vulnerable services and processes.

\paragraph{Observations.}\label{sec:obs_space} Observations are transient and only contain the outcome of the executed action, concatenated with an encoding of the action type and four mutually exclusive status flags (success, connection error, permission error, undefined error).

\paragraph{Reward.} Each action incurs a cost, and the reward $r_t$ is defined as $-\text{cost}(a_t)$ plus any additional bonus earned. For a successful subnet scan, a discovery bonus $V_d$ is added for each newly discovered host $n_{dh}$, yielding $r_t = -\text{cost}(a_t) + (V_d \times n_{dh})$. For a successful privilege escalation, a host value bonus $V_h$ is added, giving $r_t = -\text{cost}(a_t) + V_h$. In all other cases, the reward is simply $r_t = -\text{cost}(a_t)$.

\subsection{Network Generation}\label{sec:network_generation}
\begin{figure}[t]
    \centering
    \begin{subfigure}[t]{0.37\linewidth}
        \centering
        \includegraphics[height=4.5cm]{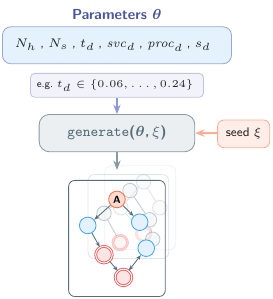}
    \end{subfigure}
    \hfill
    \begin{subfigure}[t]{0.60\linewidth}
        \centering
        \includegraphics[height=4.5cm]{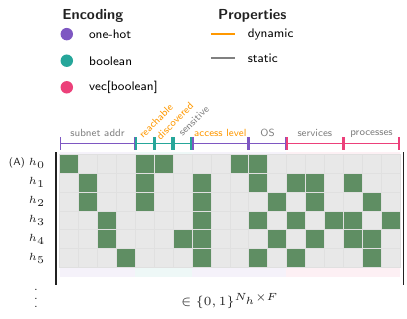}
    \end{subfigure}
    \caption{\textbf{Left:} Number of hosts ($N_h$), and subnets ($N_s$), topology ($t_d$), service ($svc_d$), process ($proc_d$) and sensitive host density ($s_d$) are all parameters that influence the generated networks. The blue nodes represent normal hosts, the red node labelled A represents the attacker's position, and the double circled red nodes represent sensitive hosts. \textbf{Right:} Illustration of the batched state representation. $F$ is the dimension of the host features.}
    \label{fig:context_gen_batched_state}
\end{figure}

Network generation proceeds in several stages and is designed to balance realism, diversity, and guaranteed solvability of the resulting scenarios. Each generated network represents a distinct learning context while adhering to structural constraints that ensure meaningful penetration testing tasks. Figure \ref{fig:context_gen_batched_state} provides an illustration for this process.

We begin by randomly distributing a fixed number of hosts $N_h$ across a fixed number of subnets $N_s$. Among these subnets, two are assigned special semantic roles that remain fixed across all generated scenarios. One subnet represents the Internet and contains a single host corresponding to the attacker’s machine. A second subnet represents a demilitarized zone (DMZ), which serves as a buffer between external and internal infrastructure. All remaining subnets are treated as arbitrary internal subnets.

Host-level properties are then assigned. Each host is randomly assigned an operating system according to a fixed distribution. Vulnerable services and processes are distributed independently across hosts using the service density $svc_d$ and process density $proc_d$ parameters, respectively. These parameters control the average number of exploitable services and processes present in the network. Host sensitivity is assigned independently with probability $s_d$. This mechanism allows the generator to model varying levels of organizational security maturity, including misconfigurations or users not following security best practices. 

To ensure that every generated scenario is solvable and supports meaningful agent interaction, we enforce several feasibility constraints after the initial random assignment. First, every subnet is guaranteed to contain at least one host running at least one service, ensuring that the agent can pivot into every subnet that becomes reachable. Second, every sensitive host is guaranteed to run at least one service and at least one process, which ensures that privilege escalation is possible on all sensitive machines. If any of these conditions are violated, the minimal number of required services or processes is added randomly to the affected hosts. These post-generation checks guarantee that successful episodes are achievable without constraining the overall diversity of generated networks. They also ensure that solve rates are comparable across $t_d$ and $N_h$.

After host and subnet properties are fixed, we generate the network topology. Connectivity between subnets is represented by an adjacency matrix of size $N_s \times N_s$, where entries are sampled according to the topology density parameter $t_d$, which controls the probability that two subnets are connected. To reflect basic network security restrictions, we impose that the Internet subnet may only communicate with the DMZ. To ensure that hosts within the same subnet can always communicate, the diagonal of the adjacency matrix is populated with ones.

The topology matrix is intentionally not required to be symmetric, so subnets may be unreachable from the attacker's starting position and effectively become inactive for an episode. This induces variation in active host counts and effective horizon length even at fixed $N_h$, providing a natural difficulty curriculum without explicitly staging scenarios.

Although the total number of hosts $N_h$ is fixed in order to allocate static memory layouts required for JIT compilation, the number of active hosts -- those belonging to reachable subnets -- varies across generated networks. As a result, a distribution over effective problem sizes is obtained even when $N_h$ is held constant. This property enables curriculum learning by exposing agents to scenarios of increasing complexity during training. We illustrate this variability empirically in Section \ref{sec:environment_configurations}.
 
\subsection{Context Sets}\label{sec:context_sets}
The design of the environment allows for several context sources. One approach is to simply use a seed, which then influences the random number generators used during the network generation phase. Train and evaluation phases can as such be seeded differently ~\citep{cobbe2019quantifying}. Further, generation parameters may be changed, to create denser or more sparse networks. The process and service density in hosts, describing how many of them are vulnerable, can also be adjusted. Changing these parameters provides the opportunity to evaluate on out-of-distribution networks, and assess zero-shot transfer capabilities of policies as discussed in \citep{kirk_survey_2023}. Importantly, the context $c$ is not included in the agent's observations; the agent must act under uncertainty about the current network configuration, inferring relevant information through interaction. This corresponds to the unobserved-context case of the CMDP as described by \citet{kirk_survey_2023}.

\section{Methodology}\label{sec:methodology}

We use PPO as our primary algorithm, the de facto choice for RL-based penetration testing~\citep{li_eppta_2023, terranova2024leveraging, simon2025learningrobustpenetrationtestingpolicies}. We additionally evaluate PQN~\citep{gallici2025simplifying}, a parallelized Q-learning variant whose GPU-native design aligns with our throughput-oriented framework. Both build on pure JAX implementations~\citep{lu2022discovered, gallici2025simplifying}. To test whether our findings depend on the feed-forward policy class, we additionally evaluate a recurrent variant, PPO-LSTM, in the zero-shot transfer experiment of Section~\ref{sec:exp_zspt}.

\paragraph{Training setup.} To cope with partial observability, we follow~\citet{simon2025learningrobustpenetrationtestingpolicies} and append the accumulated history of past observations to the current one -- mirroring how human testers track prior scan results. To avoid wasting budget on invalid actions, we apply action masking~\citep{huang_closer_2022}: we mask actions targeting undiscovered and unreachable hosts, and exploits/privilege escalations incompatible with a host's OS--service or OS--process combination. We deliberately do \emph{not} mask actions invalid due to missing privilege levels, as this signal is recoverable from the observations (Section~\ref{sec:obs_space}). The recurrent variant keeps this augmentation and adds an LSTM layer on top of it, so the two mechanisms for handling partial observability are combined rather than substituted.

\paragraph{Reward scaling.}\label{sec:reward_scaling} Since networks are procedurally generated, the number of sensitive hosts ($N_{sh}$) varies across contexts, causing high variance in episodic returns that can destabilize value estimation~\citep{van_hasselt2016learning, cobbe_leveraging_2020}. Ensuring guaranteed solvability of scenarios (Section~\ref{sec:network_generation}) allows us to scale rewards by the theoretical maximum return: $\hat{r}_t = r_t / (N_{sh} \cdot V_h)$, where $V_h$ is the sensitive-host reward minus the action cost to get there. This makes the learning signal reflect structural difficulty (vulnerability sparsity, sensitive-host count) rather than network size, yielding a comparable signal across the context distribution. This scaling is also a prerequisite for PLR: without it, regret estimates would be biased toward larger networks that accumulate higher raw returns, conflating size with learning potential.

\paragraph{Unsupervised Environment Design Methods.}\label{sec:ued_methods} Prioritized Level Replay (PLR)~\citep{jiang2021plr} works in conjunction with procedurally generated environments, and aims to form a natural curriculum of levels that are still deemed \textit{interesting} for the learning progression. These can be levels where the agent currently exhibits the highest regret with respect to its current policy. PLR performs gradient updates on the random levels, whereas robust PLR (denoted as $\text{PLR}^\bot$) only updates the agent on the levels sampled from the buffer~\citep{jiang_replay-guided_2021}. ACCEL~\cite{parker-holder_2023_accel} is an evolutionary method based on PLR. It incrementally alters the complexity of levels in the buffer through small mutations. We provide five mutations covering services, processes, OS, sensitive-host placement, and topology edges; details and post-mutation solvability checks are in Appendix~\ref{app:mutations}. We compare the different UED methods against Domain Randomization~\citep{tobin_2017_DR}, where at each end of an episode, a new network is generated from the parameterized distribution (cf. Section \ref{sec:network_generation}). We use the implementation provided by the JaxUED library \citep{coward2024JaxUED}.

\section{Experiments}\label{sec:experiments}
Our experiments investigate three aspects of NASimJax: (1) its throughput advantage and memory footprint, (2) how standard RL algorithms fare on this benchmark, and (3) how training distribution affects zero-shot generalization. Appendix~\ref{app:hyperparams} provides sweep ranges, selection protocol, and final hyperparameters. Appendix~\ref{app:computation_resources} lists the computational resources.

\subsection{Throughput and Memory Scaling}\label{sec:perf_comp}
\begin{figure}[t]
    \centering
    \begin{subfigure}[t]{0.42\linewidth}
        \centering
        \includegraphics[width=\linewidth]{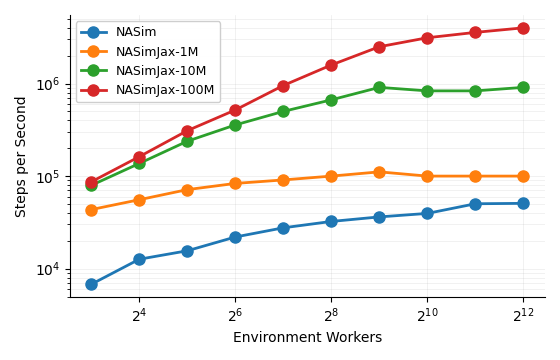}    
    \end{subfigure}
    \hfill
    \begin{subfigure}[t]{0.42\linewidth}
        \centering
        \includegraphics[width=\linewidth]{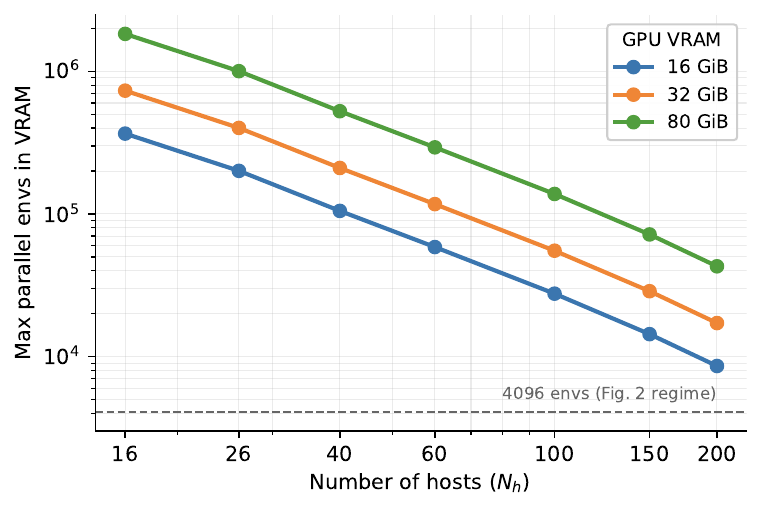}
    \end{subfigure}
    \caption{\textbf{Left:} Training speed comparison between NASim with 10M steps of training budget against NASimJax with 1M, 10M and 100M steps. Number of environment workers are doubled every time. Results show the impact of JAX's JIT-compilation on total runtime. Full results are in Appendix \ref{app:speed_comp_details}. \textbf{Right:} NASimJax's memory footprint, given an increased number of hosts.}
    \label{fig:speed_comparison_memory_profile}
\end{figure}

We compare throughput between NASim's Python implementation and NASimJax by training PPO policies with increasing numbers of parallel environment workers (8 to 4096, doubling). StochNASim and NASimEmu are CPU-bound Python implementations inheriting NASim's throughput profile; we therefore compare against NASim directly as the shared substrate. We train actual policies rather than comparing random rollouts, to reflect end-to-end usage. For NASim and NASimJax, we use the PPO implementations from \texttt{cleanRL}~\citep{huang2022cleanrl} and \texttt{PureJaxRL}~\citep{lu2022discovered}, respectively, with matched minibatch and rollout sizes. Figure~\ref{fig:speed_comparison_memory_profile} (left) shows that NASimJax reaches $4$M steps/s at 4096 workers with a 100M-step budget -- an $80\times$ speed-up over NASim. The gap widens with training budget because JIT compilation overhead amortizes over longer runs, making NASimJax especially suited to large-scale experiments. CyberBattleSim does not support parallel workers out of the box and was excluded. Figure~\ref{fig:speed_comparison_memory_profile} (right) reports the memory frontier: at 4096 parallel environments (the throughput regime above), networks of up to 200 hosts fit comfortably on 16GB of VRAM, and larger networks remain feasible at reduced parallelism. Together, the two panels show what scale of experiment is achievable with NASimJax under typical GPU budgets.

\subsection{Environment Configurations}\label{sec:environment_configurations}

Rather than prescribing fixed benchmark scenarios, NASimJax is designed as a configurable research framework. The network generation pipeline exposes a set of parameters: topology density ($t_d$), total host count ($N_h$), service ($svc_d$) and process ($proc_d$) density, sensitive host density ($s_d$) (cf. Section~\ref{sec:network_generation}). These allow researchers to instantiate environments aligned with their specific research interests. To ground the experimental investigation that follows, we describe three reference configurations of increasing complexity that we use throughout this work in Table~\ref{tab:env_config}. We emphasize that these are illustrative rather than normative. These scales exceed those typically used in prior simulation-based work (e.g. up to 8 hosts in StochNASim and 16 in PenGym and NASimEmu) though these environments differ in scope and formulation.


\begin{table}[htbp]
\centering
\caption{Network configuration and topology densities per network size. All configurations use 3 processes, 3 services, 2 operating systems, and $svc_d = proc_d = 0.7$. Training densities are \textbf{bold}.}
\label{tab:env_config}
\begin{tabular}{lcccccccccc}
\toprule
Hosts & $N_s$ & $s_d$ & Steps & $|\mathcal{A}|$ & \multicolumn{5}{c}{Topology Density $t_d$} \\
\midrule
16 & 7  & 0.20 & 300 & 256 & \textbf{0.06} & 0.115 & \textbf{0.15} & 0.195 & \textbf{0.24} \\
26 & 10 & 0.15 & 300 & 416 & \textbf{0.04} & 0.08  & \textbf{0.12} & 0.16  & \textbf{0.20} \\
40 & 16 & 0.15 & 500 & 640 & \textbf{0.03} & 0.06  & \textbf{0.09} & 0.12  & \textbf{0.15} \\
\bottomrule
\end{tabular}
\end{table}

Across all configurations, action costs are fixed at 1 for scanning and 3 for exploit and privilege escalation actions. This 3:1 ratio aims to discourage brute-force policies that skip information gathering. The sensitive host reward is $V_h = 50$ and we provide a host discovery reward of $V_d = 1$. Compromising a sensitive host strongly dominates the action cost signal, while the small discovery reward provides a mild incentive for exploration. Since rewards are scaled by $N_s \cdot V_h$ during training (cf. Section~\ref{sec:reward_scaling}), the absolute magnitude of $V_h$ does not affect the normalized learning signal -- only the relative ratios between these values matter. 

Parameters such as $t_d$ and $s_d$ were carefully chosen to enable the generation of networks with a varied number of \textit{active} hosts, as illustrated in Figure~\ref{fig:active_host_distr} (see Appendix~\ref{app:active_host_distributions} for complete distributions). The $t_d$ ranges are scaled inversely with network size because larger networks require lower connectivity probabilities to produce meaningful variation in active host counts. At higher densities, most subnets become reachable and the distribution concentrates near the maximum.

\begin{figure}[!htbp]
    \centering
    \begin{subfigure}[t]{0.3\linewidth}
        \centering
        \includegraphics[trim=1mm 1mm 1mm 1mm, clip, width=\linewidth]{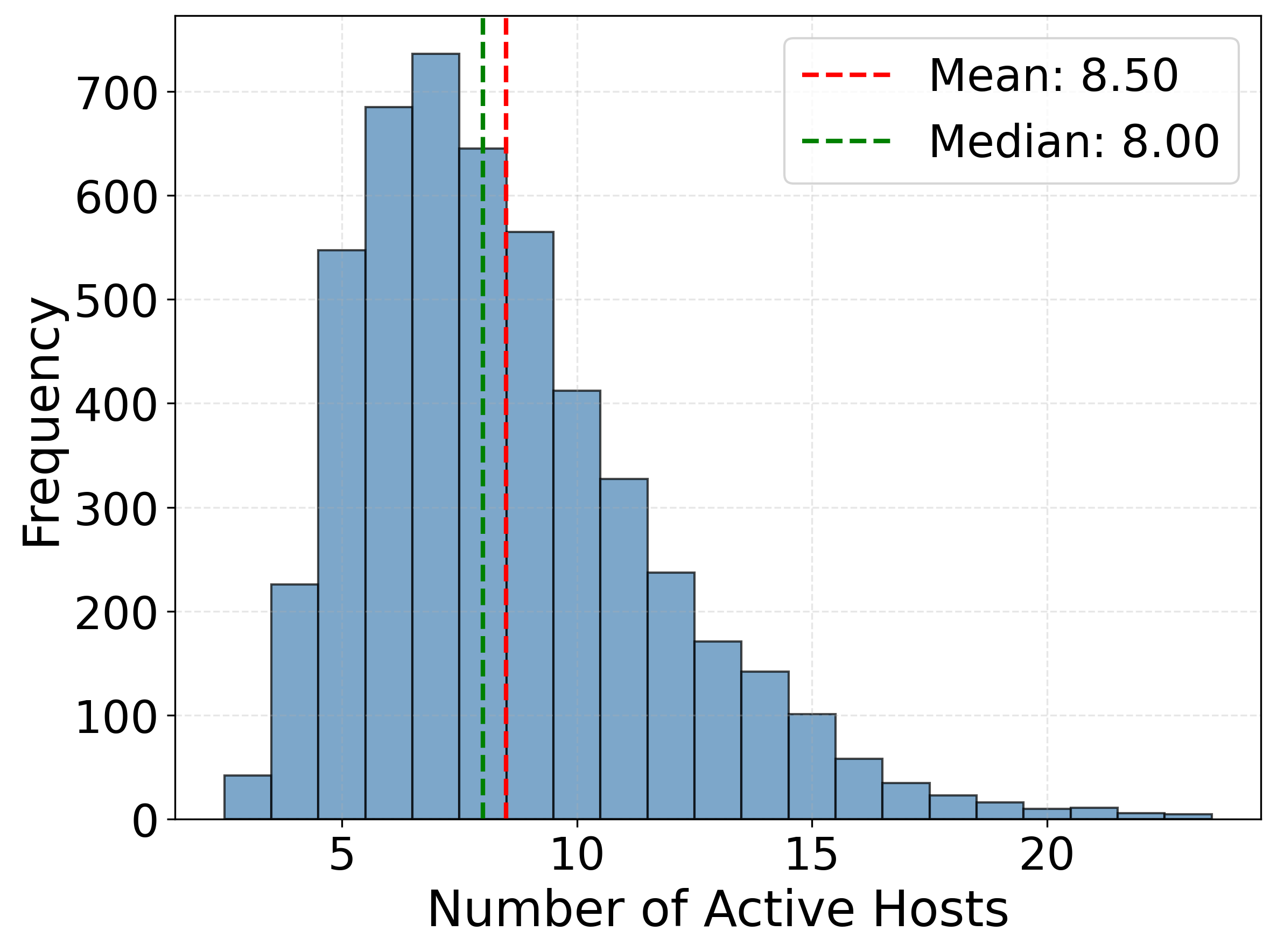}
        \caption{$t_d=0.04$}
        \label{fig:ah_distr_26h_0.04td}
    \end{subfigure}
    \hfill
    \begin{subfigure}[t]{0.3\linewidth}
        \centering
        \includegraphics[trim=1mm 1mm 1mm 1mm, clip, width=\linewidth]{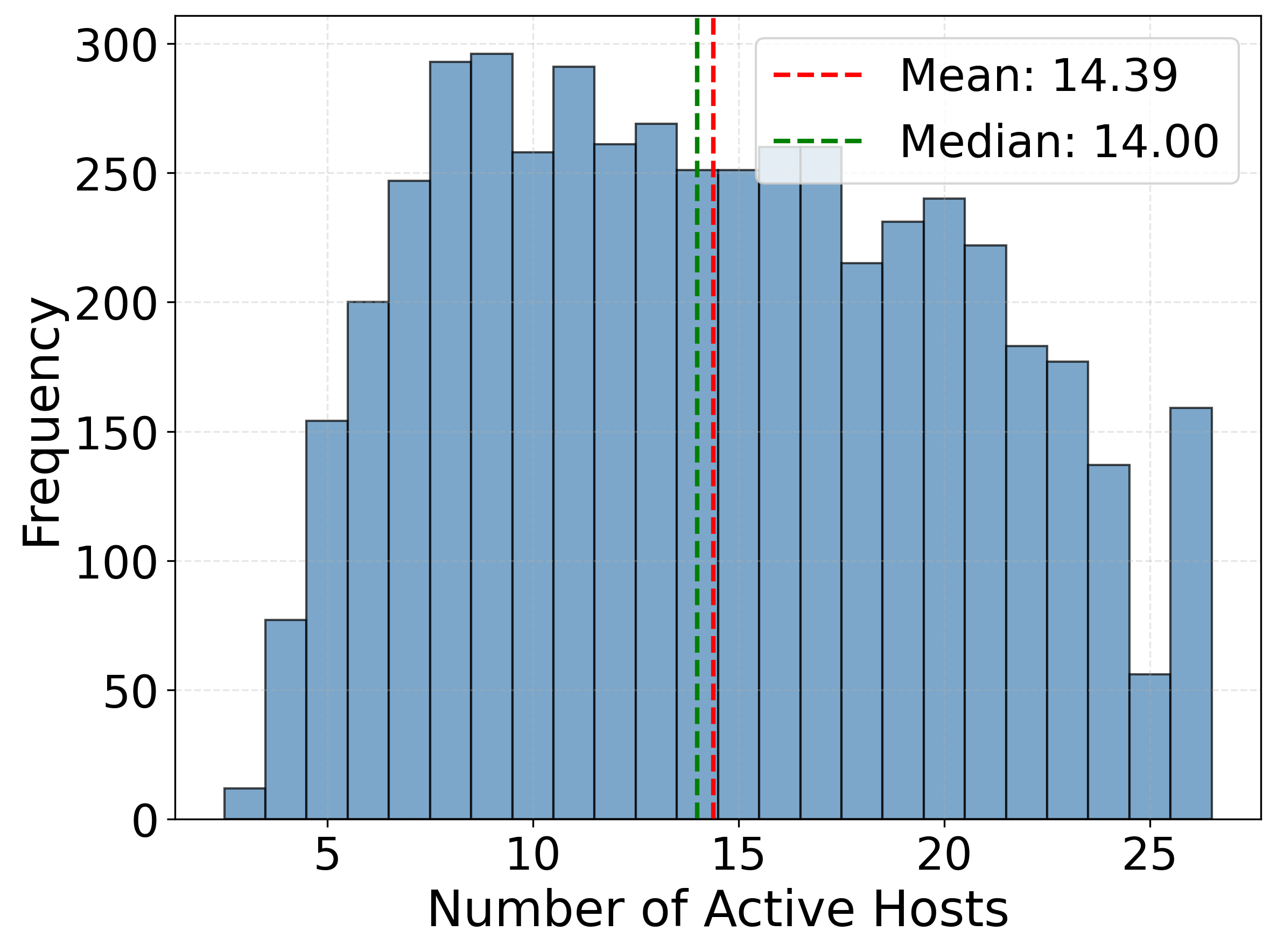}
        \caption{$t_d=0.12$}
        \label{fig:ah_distr_26h_0.12td}
    \end{subfigure}
    \hfill
    \begin{subfigure}[t]{0.3\linewidth}
        \centering
        \includegraphics[trim=1mm 1mm 1mm 1mm, clip, width=\linewidth]{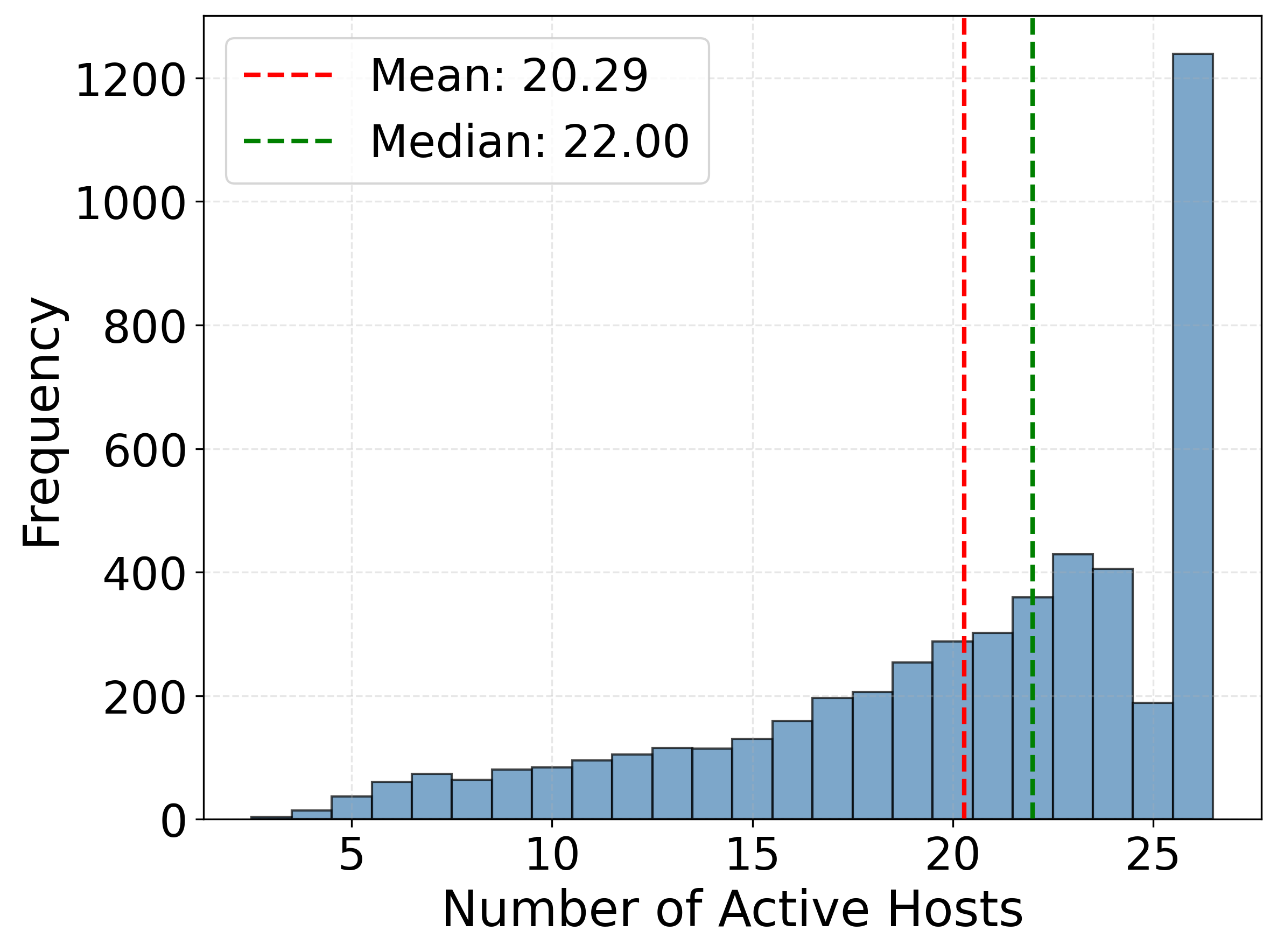}
        \caption{$t_d=0.2$}
        \label{fig:ah_distr_26h_0.2td}
    \end{subfigure}
    \caption{Visualization of $t_d$'s effect on active host counts within generated networks of 26 hosts.}
    \label{fig:active_host_distr}
\end{figure}

\subsection{Baselines}\label{sec:baselines}
\begin{figure}[t]
    \centering
    \begin{subfigure}[t]{0.30\linewidth}
        \centering
        \includegraphics[trim=1mm 1mm 1mm 1mm, clip, width=\linewidth]{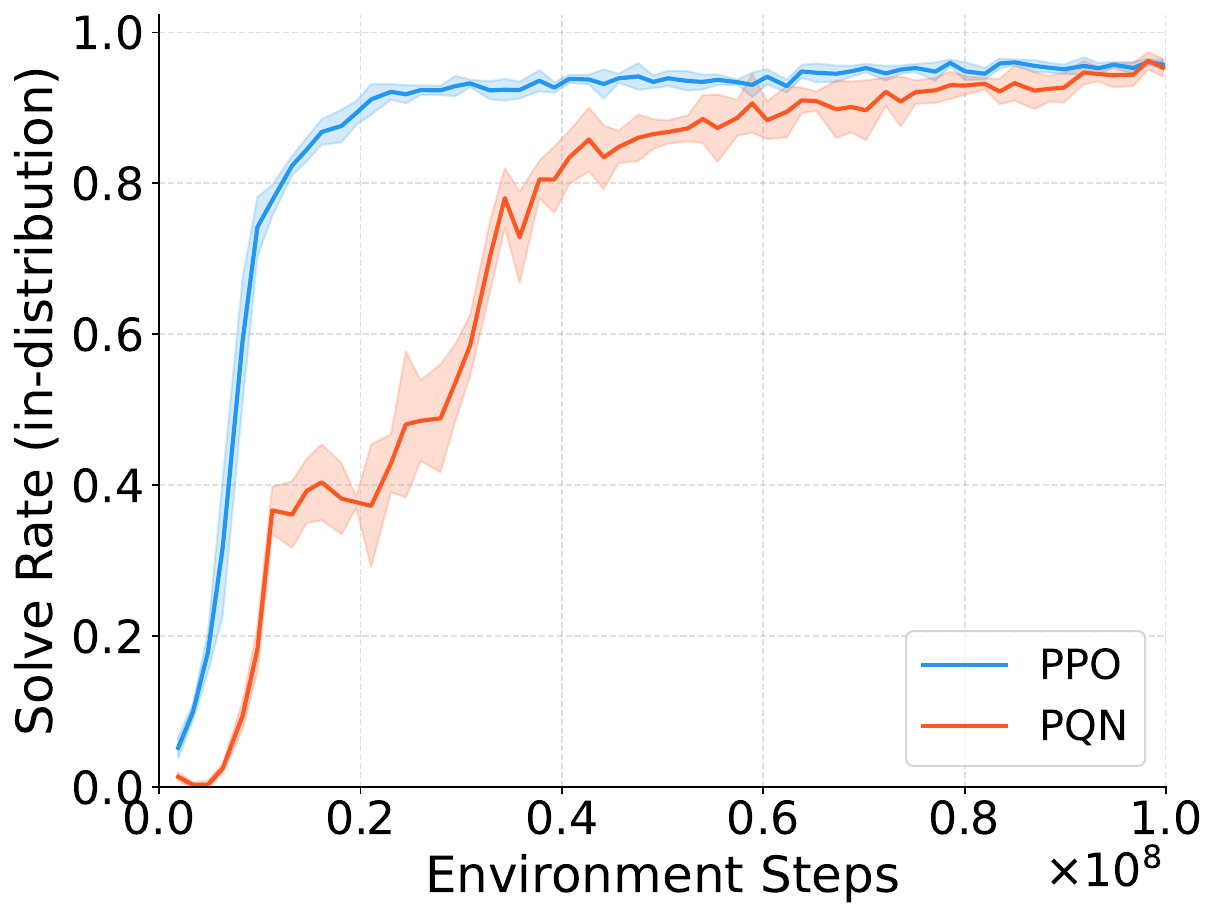}
    \end{subfigure}
    \hfill
    \begin{subfigure}[t]{0.30\linewidth}
        \centering
        \includegraphics[trim=1mm 1mm 1mm 1mm, clip, width=\linewidth]{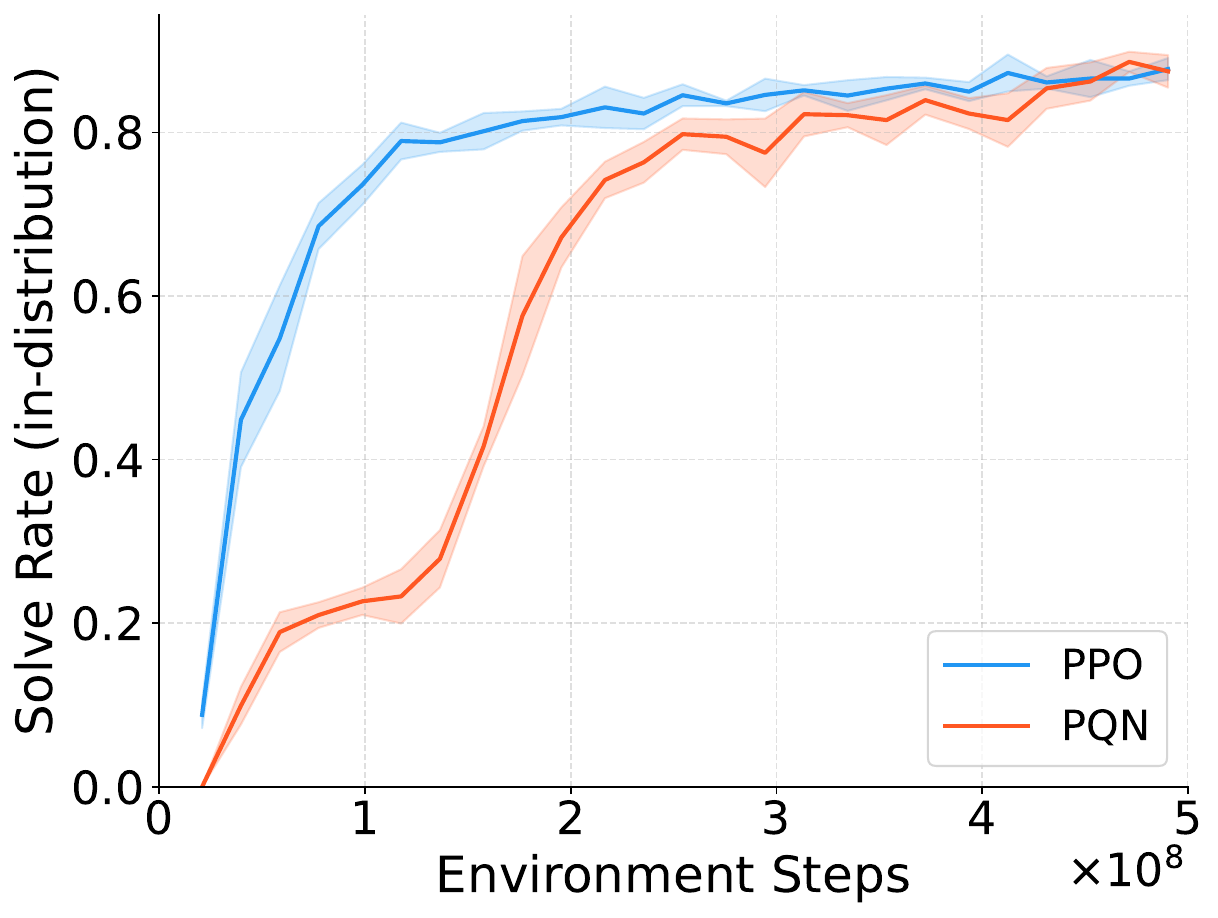}
    \end{subfigure}
    \hfill
    \begin{subfigure}[t]{0.30\linewidth}
        \centering
        \includegraphics[trim=1mm 1mm 1mm 1mm, clip, width=\linewidth]{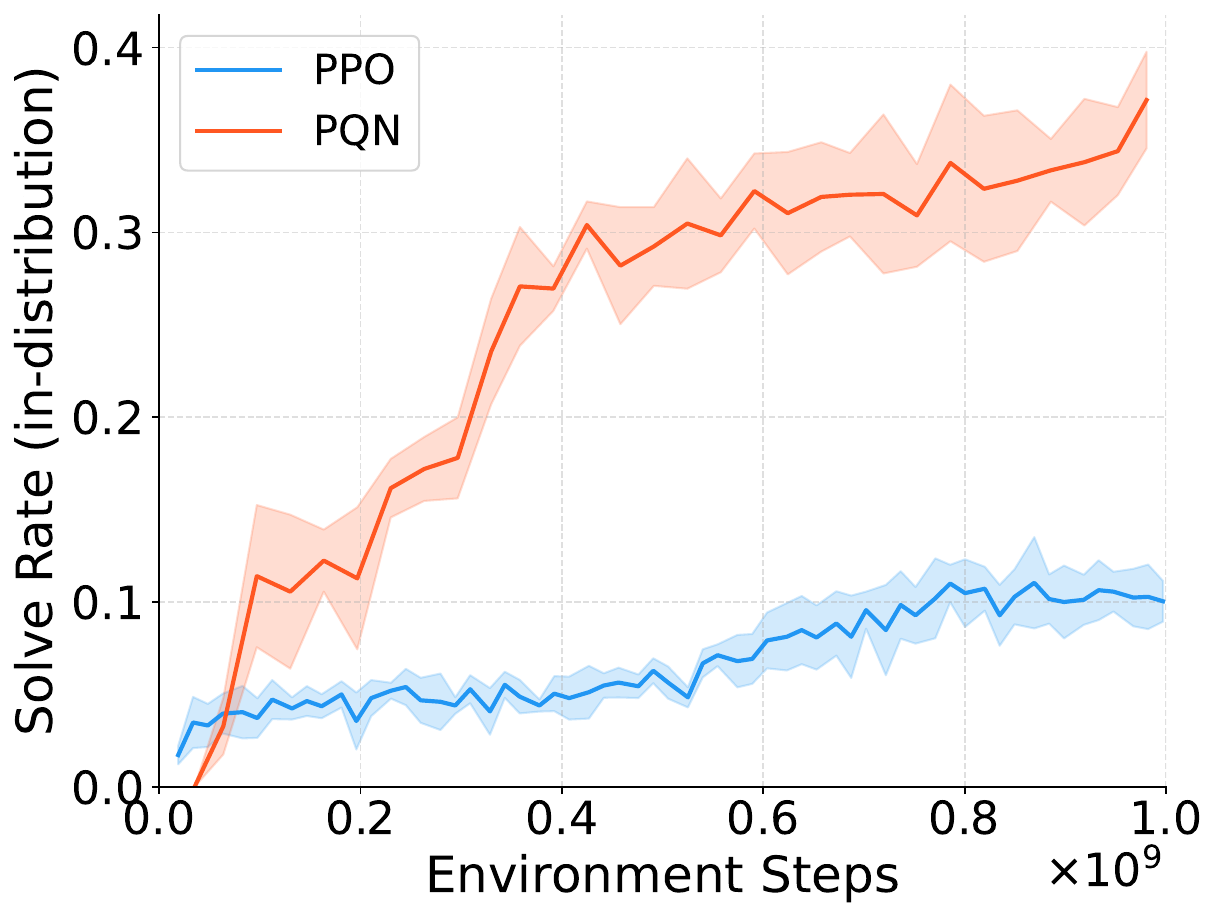}
    \end{subfigure}
    \caption{Comparison PPO and PQN on networks with \textbf{(Left)} 16 Hosts, $t_d=0.15$; \textbf{(Middle)} 26 Hosts, $t_d=0.12$; and \textbf{(Right)} 40 Hosts, $t_d=0.09$. Both algorithms use DR and evaluate in-distribution performance on 50 networks. Results are reported over 5 seeds with 95\% CI.}
    \label{fig:ppo_pqn_baselines}
\end{figure}

To establish baseline performances, we train and evaluate PPO and PQN on the configurations established in Section \ref{sec:environment_configurations} using different training budgets: 100M steps for 16-host networks, 500M for 26 hosts, and 1B for 40 hosts. Each algorithm is trained on five seeds to account for variance. We periodically evaluate the policies in-distribution over 50 networks (using the same $t_d$), as our aim here is to provide baseline performances, not measure out-of-distribution generalization. While the final solve rate is similar for 16 and 26 hosts, the results in Figure~\ref{fig:ppo_pqn_baselines} show a large difference for 40 hosts. PQN achieves an average solve rate of $\sim$37\%, compared to PPO's $\sim$15\%. Although this difference is noticeable, we'd also like to highlight that, given the low solve rate, neither of them constitutes a meaningful policy. This showcases that further efforts are still required in this domain.

\subsection{Zero-Shot Policy Transfer}\label{sec:exp_zspt}
The CMDP formulation allows us to vary the training distribution over active hosts via the $t_d$ parameter. This experiment tests how the choice of training distribution affects generalisation to unseen active host distributions at evaluation time. We use PPO and PPO-LSTM for this experiment as PLR and ACCEL are designed around on-policy regret estimation. For every network described in Table~\ref{tab:env_config}, we select five topology densities per network size to create varying distributions (cf.\ Figure~\ref{fig:active_host_distr}). The UED algorithms used are DR, PLR, $\text{PLR}^\bot$ and ACCEL (cf.\ Section~\ref{sec:ued_methods}). 

For each algorithm and network size, we train three policies, each on one of the three training densities highlighted in Table~\ref{tab:env_config}. The training budget remains the same as in our previous experiment. Each policy is then evaluated on all five densities, using one set of 50 evaluation environments per $t_d$. As a result, each policy is evaluated on four densities it did not train on, two of which (the non-bold entries in Table~\ref{tab:env_config}) are held out entirely and never used for training at that network size.


\begin{figure}[]
    \centering
    \begin{subfigure}[t]{0.3\linewidth}
        \centering
        \includegraphics[trim=1mm 1mm 1mm 1mm, clip, width=\linewidth]{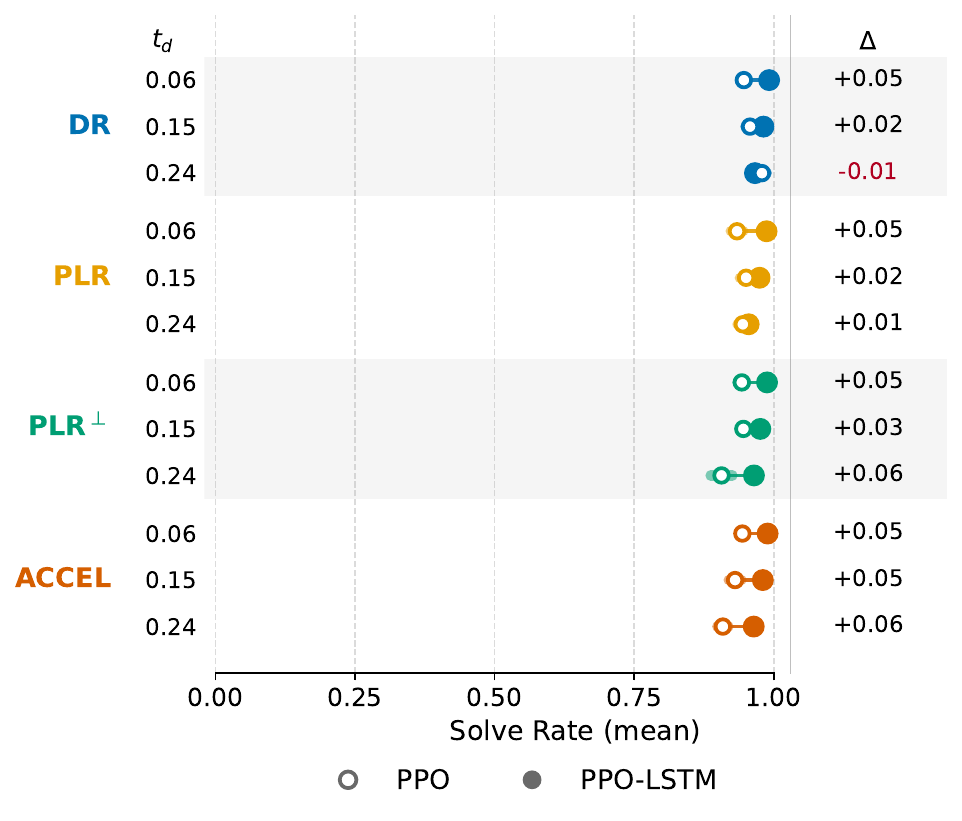}
        \caption{16 Hosts}
        \label{fig:zspt_16_hosts}
    \end{subfigure}
    \hfill
    \begin{subfigure}[t]{0.3\linewidth}
        \centering
        \includegraphics[trim=1mm 1mm 1mm 1mm, clip, width=\linewidth]{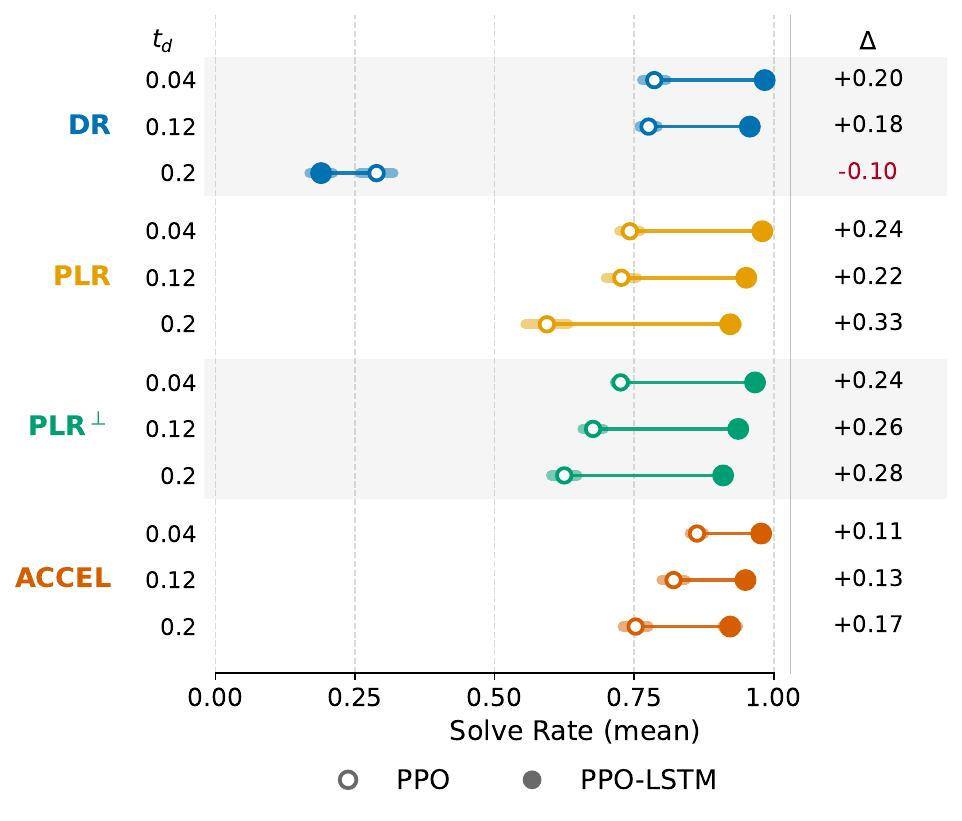}
        \caption{26 Hosts}
        \label{fig:zspt_26_hosts}
    \end{subfigure}
    \hfill
    \begin{subfigure}[t]{0.3\linewidth}
        \centering
        \includegraphics[trim=1mm 1mm 1mm 1mm, clip, width=\linewidth]{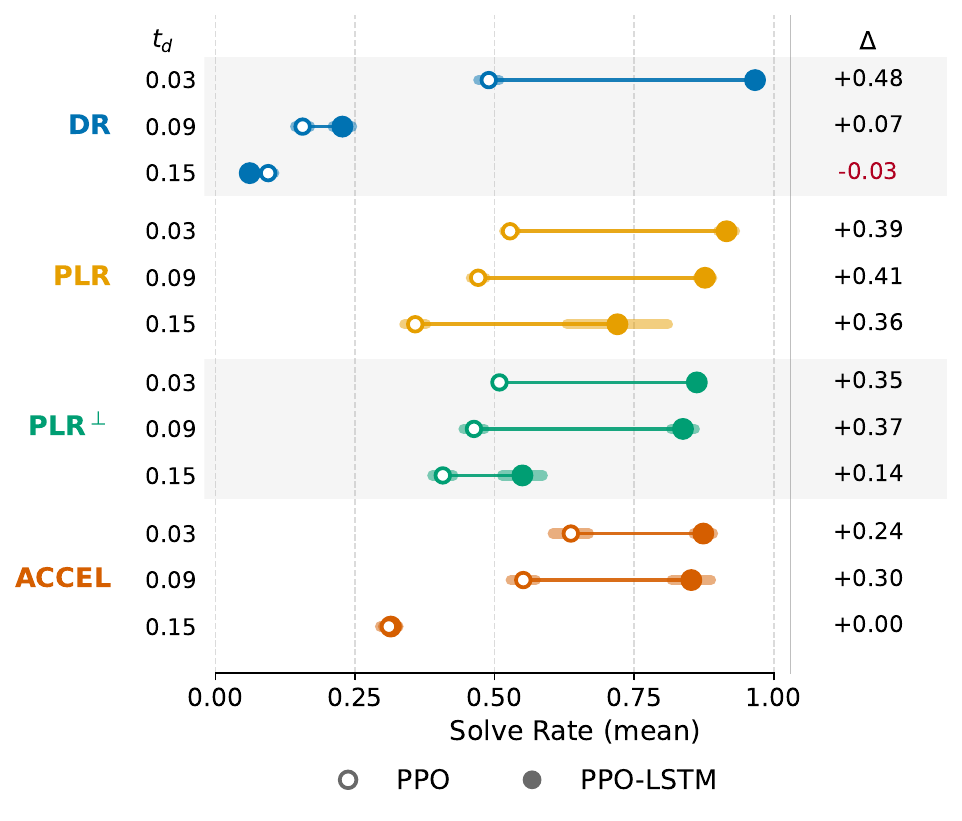}
        \caption{40 Hosts}
        \label{fig:zspt_40_hosts}
    \end{subfigure}
    \caption{ZSPT across topology densities $t_d$ for 16-, 26- and 40-host networks, PPO (open markers) vs.\ PPO-LSTM (filled). Each algorithm is trained at three values of $t_d$ (low, mid, high) and evaluated on all $t_d$. Markers give the solve rate averaged over the evaluation densities and over five seeds. The connector length is the recurrent gain, repeated numerically as $\Delta$ (red where recurrence does not help); shaded bands span $\pm1$ s.d.\ across seeds.}
    \label{fig:topo_density_dumbbell}
\end{figure}

Figure~\ref{fig:topo_density_dumbbell} shows the aggregated results; For 16-host networks, neither the algorithm nor the training distribution makes a meaningful difference at that scale. For 26- and 40-host networks, training on the lowest $t_d$ consistently yields the best evaluation performance across all densities, including those denser than the training distribution, and this holds for every algorithm under both architectures. Training at high $t_d$ degrades all methods, but most severely DR, which averages only $\sim$29\% (PPO) and $\sim$19\% (PPO-LSTM) across evaluation densities at $t_d=0.2$ on 26-host networks, where PLR, $\text{PLR}^\bot$ and ACCEL all stay above 90\% under recurrence. On 40-host networks, ACCEL leads most settings under PPO but falls behind PLR and $\text{PLR}^\bot$ at $t_d=0.15$, suggesting that mutation-based curriculum struggles when most encountered networks are already difficult; under PPO-LSTM that deficit widens from $\sim$5 to $\sim$40 points ($\sim$31\% vs.\ $\sim$72\% and $\sim$55\%).

Adding an LSTM block raises solve rates at almost every setting, without changing the picture this experiment paints. The exception is the densest training density, where recurrence brings no gain (40 hosts, $t_d=0.15$) or a small loss (26 hosts, $t_d=0.2$), confirming that DR's collapse under dense training is a property of the training distribution rather than a limitation of the policy class. Among the stronger methods the ordering moves by a few points: PLR edges ahead of ACCEL on 40-host networks, and the top three sit within roughly a point of each other at the sparser 26-host densities. The two effects the experiment surfaces remain unchanged: sparse-topology training generalises best at every size and under both architectures, and regret-based replay is what keeps a method robust when training is dense.

\section{Discussion \& Limitations}\label{sec:discussion}

\textbf{Algorithmic gap at scale.} PQN's advantage over PPO emerges only at 40 hosts (Section~\ref{sec:baselines}). This crossover is consistent with PQN's distinguishing property: LayerNorm and $\ell_2$-regularised TD admits formal stability guarantees in the long-horizon, off-policy regime that PPO's policy-gradient critic does not~\citep{gallici2025simplifying}, and its $\lambda$-returns plausibly aid credit assignment as effective horizons and action spaces grow with network size. Whether stability, multi-step credit assignment, or representational capacity dominates remains an empirical question that NASimJax is well-suited to investigate.

\textbf{Training context and zero-shot transfer.} Curriculum formation is the dominant factor in our results. Sparse-topology training produces an implicit curriculum. The agents meet shorter-horizon tasks early and build competence before facing complex attack paths. Regret-based replay produces an explicit version of the same effect, which is why PLR-family methods stay competitive at high $t_d$ where DR collapses. ACCEL's sensitivity to initial level difficulty is an expected consequence of its evolutionary design~\citep{parker-holder_2023_accel}: small mutations extend the curriculum outward from achievable levels, but cannot manufacture them when the generator predominantly produces hard instances, as at $t_d=0.15$ on 40 hosts. Practically, where the deployment distribution is dense, training on cheaper sparse networks is preferable to training on-distribution.

\textbf{Evaluative Scope \& Limitations.}
Beyond the modelling abstractions outlined in Section~\ref{sec:nasimjax}, several empirical and interpretive limitations constrain our claims.  First, algorithmic coverage spans PPO (feed-forward and recurrent) and PQN; transformer-based, permutation-invariant and off-policy actor--critic designs remain unexplored. Repeating the zero-shot transfer experiment with a recurrent policy (Section~\ref{sec:exp_zspt}) raises the absolute solve rates throughout reproduces and the mechanisms we report: DR's collapse when training is dense, and ACCEL's difficulty when the training distribution is already hard. The exact ordering among the strongest methods shifts by a few points in places, but overall findings remain intact. Second, our zero-shot transfer study varies only topology density $t_d$, holding $svc_d$, $proc_d$, $s_d$, and $N_h$ fixed within each configuration, so generalization claims apply along the topology axis rather than to arbitrary distributional shift. Third, we assess policies via solve rate on fixed held-out network sets per density, which captures task completion but not action efficiency, stealth, or attack-path quality. Finally, NASimJax supports evaluative claims about RL behaviour on procedurally generated networks under our generator: improvements should be interpreted as evidence about scalable distributional training in this abstraction. Transfer to real enterprise networks, to emulation backends such as NASimEmu or PenGym, and to defender-aware settings is left as a separate question that this framework does not by itself adjudicate.

\section{Conclusion} We introduced NASimJax, a JAX-native Contextual POMDP framework and configurable benchmark for penetration testing that resolves the throughput bottleneck of existing simulators (up to $80\times$) and exposes a controllable distribution over network instances along topology, size, and vulnerability density. Together with the limitations articulated above, our baseline and UED findings position NASimJax as a configurable, guaranteed-solvable platform for advancing RL in this domain, with capacity to support future work on action-space scaling, topology effects, architectural design, and curriculum methods at scales that prior simulators could not reach.

\begin{ack}
This research was funded by the Royal Higher Institute of Defence under the project DAP23/05. Further support comes from the Flemish Government under the “Onderzoeksprogramma Artificiële Intelligentie (AI) Vlaanderen” program. The resources and services used in this work were, in part, provided by the VSC (Flemish Supercomputer Center), funded by the Research Foundation - Flanders (FWO) and the Flemish Government. Pieter Libin acknowledges support from the Research council of the Vrije Universiteit Brussel (OZR-VUB) via grant number OZR3863BOF. We also thank Hicham Azmani, Bram Silue and all other lab members that have provided feedback throughout the creation of this work.
\end{ack}

\bibliography{bibliography}
\bibliographystyle{unsrtnat}

\appendix
\onecolumn
\input{appendix}


\end{document}

%% file: appendix.tex
\section{Contextual Reinforcement Learning Setting}
To readers more familiar with Cyber-Security than Reinforcement Learning, we provide an extended description of the notations presented in Section~\ref{sec:contextual_rl}.

\subsection{Markov Decision Process} An MDP~\citep{puterman_1990_mdp} is a tuple $\langle \mathcal{S}, \mathcal{A}, \mathcal{P}, \mathcal{R}, \rho_0, \gamma \rangle$ where $\mathcal{S}$ is the state space, $\mathcal{A}$ is the action space, $\mathcal{P}(s'|s,a)$, is the transition probability distribution over next states, conditioned on the current state and action; $\mathcal{R} : \mathcal{S} \times \mathcal{A} \times \mathcal{S}  \rightarrow \mathbb{R}$, a reward function; $\rho_0 \in \Delta(\mathcal{S})$, the initial state distribution, from which an initial state can be sampled: $s_0 \sim \rho_0 $; and $\gamma \in [0,1]$ a discount factor, that determines the relative weighting of future rewards. The objective is to learn a policy $\pi(a|s)$, a probability distribution over actions conditioned on the current state, that maximizes the expected cumulative discounted reward $\mathbb{E}_\pi[G_0]$ where $G_t = \sum_{k=0}^{\infty} \gamma^k r_{t+k}$ is called the return. $r_t$ is a random variable that represents the reward obtained at time step $t$. In finite-horizon tasks the sum is clipped to $T$, the maximum number of allowed steps.

\subsection{Partially Observable Markov Decision Process} A POMDP~\citep{sutton_rl_2018} extends the MDP framework to a tuple $\langle \mathcal{S}, \mathcal{A}, \mathcal{P}, \mathcal{R}, \Omega, \mathcal{O}, \rho_0,\gamma \rangle$, where $\Omega$ represents the observation space and $\mathcal{O}(o|s',a)$ defines the observation function. The agent does not directly observe the underlying state. Only parts of it can be observed at a given time through $\mathcal{O}$. The optimal policy now depends on the history of observations or belief state $b_t \in \Delta(\mathcal{S})$.

\subsection{Contextual Markov Decision Process} Following \citep{hallak2015contextual} and the formulation adopted in \citep{kirk_survey_2023}, a Contextual Markov Decision Process (CMDP) is a special class of POMDP with a context variable $c \in \mathcal{C}$, where $\mathcal{C}$ denotes the context space and $c$ is sampled at the beginning of each episode from a distribution $p(c)$. Conditioned on $c$, the environment induces a POMDP with context-dependent dynamics and rewards. Formally, a contextual POMDP is defined by the tuple $\langle \mathcal{S}, \mathcal{A}, \mathcal{P}_c, \mathcal{R}_c, \Omega, \mathcal{O}_c, \rho_{0,c}, \gamma, \mathcal{C}, p(c) \rangle$, where $\mathcal{P}_c(s' \mid s,a)$ and $\mathcal{R}_c(s,a)$ denote the transition and reward functions parameterized by $c$, $\mathcal{O}_c(o \mid s',a)$ is the observation function, and $\rho_{0,c}$ is the initial state distribution conditioned on $c$. The joint distribution over context and initial state factorizes as $p(c, s_0) = p(c)\,\rho_{0,c}(s_0)$. The context remains fixed throughout an episode but varies across episodes according to $p(c)$, thereby inducing a distribution over tasks. The objective is to learn a policy that maximizes expected return under the context distribution, i.e., $\mathbb{E}_{c \sim p(c)} \left[ \mathbb{E}_{\pi}\left[ G_0 \mid c \right] \right]$.

\section{Implementation Details}\label{app:implementation_details}

A note about software licenses. NASimJax is released under the MIT license. Algorithmic baselines adapted from JaxUED \citep{coward2024JaxUED} and PureJaxQL \citep{gallici2025simplifying} retain their upstream Apache-2.0 license; full per-file attribution is provided in \texttt{THIRD\_PARTY\_NOTICES.md}. The remainder of this appendix provides the full semantics of host properties and encoding details elided from Section~\ref{sec:task-spec}.

\subsection{State Representation}\label{app:state_repr}

\paragraph{Host properties.} Each host carries the following attributes:
\begin{itemize}
  \item \texttt{subnet address}: identifies the subnet the host belongs to (one-hot encoded).
  \item \texttt{reachable}: boolean flag, true once the host is network-reachable from the attacker.
  \item \texttt{discovered}: boolean flag, true once the host has been observed via subnet scan.
  \item \texttt{access level}: current privilege level held by the agent on the host (one-hot encoded over \{none, user, root\}).
  \item \texttt{OS}: the host's operating system (one-hot encoded).
  \item \texttt{services}: vector of boolean flags indicating which vulnerable services are running.
  \item \texttt{processes}: vector of boolean flags indicating which vulnerable processes are running.
\end{itemize}

\paragraph{Static vs. dynamic properties.} During episode execution, only three host values can change: \texttt{reachable}, \texttt{discovered}, and \texttt{access level}. All other properties are fixed during network generation and remain static throughout the episode. This separation allows the transition function to be implemented as targeted bitwise updates on a small subset of the state tensor, which is one of the factors contributing to the throughput gains reported in Section~\ref{sec:perf_comp}.

\paragraph{Services and processes as filtered abstractions.} The \texttt{services} and \texttt{processes} attributes encode only those services and processes that are vulnerable and therefore exploitable by the agent. They do not represent the complete software inventory of a host, but rather a filtered abstraction capturing actionable attack surfaces. As a consequence, hosts may have zero vulnerable services or processes, modeling systems that are correctly configured or hardened. This design keeps the state representation compact while preserving the combinatorial structure of vulnerability-driven exploitation, which is the principal driver of action-space growth in Equation~\ref{eq:action-space}.

\paragraph{Adjacency matrix and firewall rules.} Inter-subnet connectivity is represented by an $N_s \times N_s$ adjacency matrix with an additional dimension accommodating per-service firewall rules, which determine whether a given service is permitted to traverse a subnet boundary. The diagonal is populated with ones so that intra-subnet communication is always permitted. The matrix is not required to be symmetric, which allows the generator to produce directional connectivity patterns (see Section~\ref{sec:network_generation}).

\paragraph{Encoding rationale and throughput impact.} Consolidating host properties into a single $\{0,1\}^{N_h \times F}$ tensor (Figure~\ref{fig:context_gen_batched_state}) with unsigned-integer storage (one byte per property) has two advantages. First, the majority of transition logic reduces to bitwise operations, which carry very small computational overhead under JIT compilation. Second, state properties can be stored using unsigned integers that only use one byte in memory, reducing the overall memory footprint of the environment, therefore allowing for more parallel environments during training.

\paragraph{Homogeneous Subnets.} In addition to fully random host configurations, we support the generation of homogeneous subnets to better reflect real-world network administration practices, where hosts within the same subnet are often configured with similar software stacks. In this mode, services and processes are sampled jointly for all hosts in a subnet from Beta distributions parameterized by $svc_d$ and $proc_d$. This approach allows controlled variation in intra-subnet diversity while preserving the overall density of services and processes.


\subsection{Mutations}\label{app:mutations}

We provide five mutations for generated networks: 
\begin{enumerate}
    \item Toggle service. Adds or removes a service from a random host with equal probability.
    \item Toggle process. Adds or removes a process from a random host with equal probability.
    \item Change OS. Performs an OS change to a random host.
    \item Move sensitive. Moves the sensitive flag from one host to another. This changes the goal condition of the generated network.
    \item Flip network edge. Flips the edge between two subnets.
\end{enumerate}

After mutations, we perform checks whether the solvability is still guaranteed and apply patches if necessary. This is especially crucial for the last mutation, where the sole subnet with sensitive host(s) can get disconnected. Therefore, to still provide guaranteed solvable networks, a new sensitive host that is reachable needs to be chosen that is not the starting host and within the DMZ.

\section{Speed Comparison Details}\label{app:speed_comp_details}

Table \ref{tab:nasim_performance_rtx_5090} showcases the full details of our experiment in Section \ref{sec:perf_comp}.

\begin{table*}[htbp]
\centering
\caption{Performance Comparison of NASim and NASimJax Implementations on small environment}
\label{tab:nasim_performance_rtx_5090} 
\begin{tabular}{r|cc|cc|cc|cc}
\hline
\textbf{Num} & \multicolumn{2}{c|}{\textbf{NASim}} & \multicolumn{2}{c|}{\textbf{NASimJax-1M}} & \multicolumn{2}{c|}{\textbf{NASimJax-10M}} & \multicolumn{2}{c}{\textbf{NASimJax-100M}} \\
\textbf{Envs} & \textbf{Time (s)} & \textbf{Steps/s} & \textbf{Time (s)} & \textbf{Steps/s} & \textbf{Time (s)} & \textbf{Steps/s} & \textbf{Time (s)} & \textbf{Steps/s} \\
\hline
8 & 1487 & 6,724 & 23 & 43,478 & 126 & 79,365 & 1154 & 86,655 \\
16 & 789 & 12,674 & 18 & 55,556 & 73 & 136,986 & 617 & 162,075 \\
32 & 642 & 15,576 & 14 & 71,429 & 42 & 238,095 & 323 & 309,598 \\
64 & 457 & 21,881 & 12 & 83,333 & 28 & 357,143 & 193 & 518,135 \\
128 & 358 & 27,932 & 11 & 90,909 & 20 & 500,000 & 105 & 952,381 \\
256 & 305 & 32,786 & 10 & 100,000 & 15 & 666,667 & 63 & 1,587,302 \\
512 & 272 & 36,764 & 9 & 111,111 & 11 & 909,091 & 40 & 2,500,000 \\
1024 & 250 & 40,000 & 10 & 100,000 & 12 & 833,333 & 32 & 3,125,000 \\
2048 & 195 & 51,282 & 10 & 100,000 & 12 & 833,333 & 28 & 3,571,429 \\
4096 & 197 & 51,546 & 10 & 100,000 & 11 & 909,091 & 25 & 4,000,000 \\
\hline
\end{tabular}
\end{table*}


\section{Hyperparameters}\label{app:hyperparams}
We perform a hyperparameter search using Bayesian-sampling for every discussed algorithm in Section \ref{sec:methodology}, with a budget of at lest 275 trials. The number of trials for every algorithm is listed in Table~\ref{tab:sweep-compute}. Each sampled hyperparameter set is trained on 26 host networks. The performance of a run is reported as the mean solve rate over the set of evaluation levels during the final evaluation phase. After the sweep, to select the final parameter sets, we evaluate the top five most promising hyperparamters sets using five control seeds and pick the best performing and most stable ones. This selection process is repeated for every network size.

\subsection{Hyperparameter Ranges}\label{app:hyperparam_ranges}

This section discusses the search ranges and the exact hyperparameters we've selected for the experiments in Section \ref{sec:experiments}. Some more information regarding the tuning strategy: We first tuned PPO with DR. Then based on these results, narrowed down the search space for the PLR-based methods. The ranges selected for the PLR-based methods are based on the original works of Jiang et~al.~\citep{jiang2021plr, jiang_replay-guided_2021}. We run two sweeps for both PLR and PLR$^\bot$, each using a different prioritization, to further limit the search space. This is because $top_k$ prioritization additionally needs to sample the $k$, which $rank$-based prioritization does not. For UED methods using PPO-LSTM we use used reduced search spaces based on relevant hyperparamters found through the previously executed PPO sweeps. We only thus only present the selected values for PPO-LSTM.

\begin{table}[]
\centering
\caption{Hyperparameter Search Ranges: PPO with DR}
\label{tab:search_ranges_ppo_dr}
\begin{tabular}{l l}
    \toprule
    \textbf{Hyperparameter} & \textbf{Considered Values} \\
    \midrule
    Learning Rate (LR) & \{5e-5, 1e-4, 2e-4, 3e-4, 5e-4\} \\
    \# Environments & \{1024\} \\
    \# Steps & \{16, 32, 64, 128\} \\
    Layer Size & \{256, 512\} \\
    Activation Function & \{tanh\} \\
    Discount Factor ($\gamma$) & \{0.975, 0.99, 0.995\} \\
    GAE Lambda ($\lambda$) & \{0.8, 0.9, 0.95\} \\
    Clip Epsilon & \{0.1, 0.2, 0.3\} \\
    Entropy Coef. & \{0.005, 0.01, 0.02, 0.05\} \\
    Max Gradient Norm & \{0.5, 1.0, 1.5\} \\
    Value Function Coef. & \{0.25, 0.5, 1.0\} \\
    Update Epochs & \{4\} \\
    \bottomrule
\end{tabular}
\end{table}

\begin{table}[]
\centering
\caption{Hyperparameter Search Ranges: PQN with DR}
\label{tab:search_ranges_pqn_dr}
\begin{tabular}{l l}
    \toprule
    \textbf{Hyperparameter} & \textbf{Considered Values} \\
    \midrule
    Learning Rate (LR) & \{1e-4, 2e-4, 3e-4\} \\
    \# Environments & \{1024\} \\
    \# Steps & \{32, 64, 128\} \\
    Layer Size & \{256, 512\} \\
    Activation Function & \{tanh\} \\
    Discount Factor ($\gamma$) & \{0.975, 0.99, 0.995\} \\
    GAE Lambda ($\lambda$) & \{0.3, 0.5, 0.65, 0.8, 0.9\} \\
    Epsilon Start ($\epsilon$) & \{1.0\} \\
    Epsilon Finish ($\epsilon$) & \{0.01, 0.03, 0.05, 0.08, 0.1, 0.12\} \\
    Epsilon Decay ($\epsilon$) & \{0.8, 0.5, 0.3\} \\
    \# Minibatches & \{4, 8, 16\} \\
    Update Epochs & \{4\} \\
    Reward Scale & \{1.0\} \\
    \bottomrule
\end{tabular}
\end{table}

\begin{table}[htbp]
\centering
\caption{Hyperparameter Search Ranges: PLR and $\text{PLR}^\bot$}
\label{tab:search_ranges_plr_robust_plr}
\begin{tabular}{l l}
    \toprule
    \textbf{Hyperparameter} & \textbf{Considered Values} \\
    \midrule
    Learning Rate (LR) & \{2e-4, 3e-4, 5e-4\} \\
    \# Steps & \{32, 64, 128\} \\
    Layer Size & \{256, 512\} \\
    GAE Lambda ($\lambda$) & \{0.8, 0.95\} \\
    Clip Epsilon & \{0.1, 0.2, 0.3\} \\
    Entropy Coef. & \{0.005, 0.01, 0.05\} \\
    Replay Probability ($\rho$) & \{0.5, 0.7, 0.9\} \\
    Staleness Coefficient ($\omega$) & \{0.1, 0.3, 0.5, 0.7\} \\
    Temperature ($\beta$) & \{0.1, 0.5, 1.0, 1.4, 2.0\} \\
    Level Buffer Capacity & \{5000, 10000\} \\
    Score Function & \{MaxMC, PVL\} \\
    Prioritization & \{Rank, Top-$k$\} \\
    Top-$k$ ($k$) & \{5 , 25, 50, 100\} \\
    \bottomrule
\end{tabular}
\end{table}

\begin{table}[]
\centering
\caption{Hyperparameter Search Ranges: ACCEL}
\label{tab:search_ranges_accel}
\begin{tabular}{l l}
    \toprule
    \textbf{Hyperparameter} & \textbf{Considered Values} \\
    \midrule
    Learning Rate (LR) & \{2e-4, 3e-4, 5e-4\} \\
    \# Steps & \{64, 128\} \\
    GAE Lambda ($\lambda$) & \{0.8, 0.95\} \\
    Clip Epsilon & \{0.1, 0.2, 0.3\} \\
    Entropy Coef. & \{0.005, 0.01\} \\
    Replay Probability ($\rho$) & \{0.5, 0.7, 0.8, 0.9\} \\
    Staleness Coefficient ($\omega$) & \{0.1, 0.3, 0.5, 0.7\} \\
    Temperature ($\beta$) & \{0.1, 0.3, 0.5, 1.0, 2.0\} \\
    Level Buffer Capacity & \{1000, 4000, 10000\} \\
    Score Function & \{MaxMC, PVL\} \\
    Prioritization & \{Rank\} \\
    \# Edits & \{3, 5, 10\} \\
    \bottomrule
\end{tabular}
\end{table}

\subsection{Experiment Hyperparameters}\label{app:exp_hyperparam}
The following hyperparameters were held constant across all algorithms and network  sizes: number of environments (1024), activation function (\texttt{tanh}), discount  factor (0.995), and number of update epochs (4). In addition, we fix the sizes of the minibatches to 4096. For PLR and PLR$^\bot$ variants, the level buffer capacity (10{,}000), minimum fill ratio (0.5), prioritization strategy  (rank-based), and top-$k$ (disabled) were additionally held constant.

For PQN, we fixed the reward scale at $1.0$. We also empirically tested whether greedy evaluation is better than stochastic evaluation and found that greedy evaluation works better. We ran this test because of the stochastic nature of the environment, and on the basis that generally, stochastic policies perform better in partially observable environments.

For PPO-LSTM we held constant: number of environments (1024), activation function (\texttt{tanh}), discount factor (0.995), update epochs (4), trunk layer size (512), LSTM layers (1), and LR annealing. For the PLR family: rank prioritization, top-$k$ disabled, level buffer capacity (10{,}000), and minimum fill ratio (0.5, except ACCEL at 16 hosts). \textit{Exploratory grad updates} (True for PLR and ACCEL, False for PLR$^\bot$) is the structural PLR/PLR$^\bot$ distinction, not a tuned value.

\begin{table}[]
\centering
\caption{Hyperparameters for PPO DR across 16, 26, and 40 host networks.}
\label{tab:hyperparams_dr}
\begin{tabular}{l ccc}
    \toprule
    & \multicolumn{3}{c}{\textbf{PPO DR}} \\
    \cmidrule(lr){2-4}
    \textbf{Hyperparameter} & \textbf{16h} & \textbf{26h} & \textbf{40h} \\
    \midrule
    Learning Rate           & 3e-4  & 3e-4  & 3e-4  \\
    \# Steps                & 32    & 128   & 64    \\
    Layer Size              & 512   & 512   & 512   \\
    GAE Lambda ($\lambda$)  & 0.8   & 0.8   & 0.8   \\
    Clip Epsilon            & 0.1   & 0.2   & 0.1   \\
    Entropy Coef.           & 0.005 & 0.005 & 0.005 \\
    Max Grad Norm           & 1.0   & 0.5   & 0.5   \\
    Value Coef.             & 1.0   & 0.5   & 0.25  \\
    \bottomrule
\end{tabular}
\end{table}

\begin{table}[]
\centering
\caption{Hyperparameters for PQN DR across 16, 26, and 40 host networks.}
\label{tab:hyperparams_pqn_dr}
\begin{tabular}{l ccc}
    \toprule
    & \multicolumn{3}{c}{\textbf{PQN DR}} \\
    \cmidrule(lr){2-4}
    \textbf{Hyperparameter} & \textbf{16h} & \textbf{26h} & \textbf{40h} \\
    \midrule
    Learning Rate                   & 2e-4  & 2e-4  & 2e-4  \\
    \# Steps                        & 32    & 128   & 128   \\
    Layer Size                      & 512   & 512   & 512   \\
    Discount Factor ($\gamma$)      & 0.975 & 0.99  & 0.99  \\
    GAE Lambda ($\lambda$)          & 0.3   & 0.65  & 0.65  \\
    Epsilon Start ($\epsilon$)      & 1.0   & 1.0   & 1.0   \\
    Epsilon Finish ($\epsilon$)     & 0.01  & 0.01  & 0.01  \\
    Epsilon Decay ($\epsilon$)      & 0.3   & 0.3   & 0.3   \\
    \# Minibatches                  & 8     & 8     & 8     \\
    Max Grad Norm                   & 0.5   & 0.5   & 0.5   \\
    Reward Scale                    & 1.0   & 1.0   & 1.0   \\
    
    \bottomrule
\end{tabular}
\end{table}

\begin{table}[]
\centering
\caption{Hyperparameters for PLR across 16, 26, and 40 host networks.}
\label{tab:hyperparams_plr}
\begin{tabular}{l ccc ccc}
    \toprule
    & \multicolumn{3}{c}{\textbf{PPO PLR}} \\
    \cmidrule(lr){2-4}
    \textbf{Hyperparameter} & \textbf{16h} & \textbf{26h} & \textbf{40h} \\
    \midrule
    Learning Rate           & 3e-4  & 3e-4  & 3e-4  \\
    \# Steps                & 128   & 64    & 64    \\
    Layer Size              & 512   & 512   & 512   \\
    GAE Lambda ($\lambda$)  & 0.8   & 0.8   & 0.8   \\
    Clip Epsilon            & 0.3   & 0.3   & 0.3   \\
    Entropy Coef.           & 0.01  & 0.005 & 0.01  \\
    Max Grad Norm           & 0.5   & 0.5   & 0.5   \\
    Value Coef.             & 0.5   & 0.5   & 0.5   \\
    Replay Prob.\ ($\rho$)  & 0.5   & 0.5   & 0.5   \\
    Staleness ($\omega$)    & 0.7   & 0.3   & 0.5   \\
    Temperature ($\beta$)   & 0.1   & 2.0   & 2.0   \\
    Score Function          & PVL   & MaxMC & MaxMC \\
    \bottomrule
\end{tabular}
\end{table}

\begin{table}[]
\centering
\caption{Hyperparameters for PLR$^\bot$ across 16, 26, and 40 host networks.}
\label{tab:hyperparams_robust_plr}
\begin{tabular}{l ccc ccc}
    \toprule
    & \multicolumn{3}{c}{\textbf{PPO PLR$^\bot$}} \\
    \cmidrule(lr){2-4}
    \textbf{Hyperparameter} & \textbf{16h} & \textbf{26h} & \textbf{40h} \\
    \midrule
    Learning Rate           & 3e-4  & 3e-4  & 3e-4  \\
    \# Steps                & 128   & 128   & 128   \\
    Layer Size              & 512   & 512   & 512   \\
    GAE Lambda ($\lambda$)  & 0.95  & 0.95  & 0.95  \\
    Clip Epsilon            & 0.3   & 0.3   & 0.3   \\
    Entropy Coef.           & 0.005 & 0.005 & 0.005 \\
    Max Grad Norm           & 0.5   & 0.5   & 0.5   \\
    Value Coef.             & 0.5   & 0.5   & 0.5   \\
    Replay Prob.\ ($\rho$)  & 0.9   & 0.9   & 0.9   \\
    Staleness ($\omega$)    & 0.1   & 0.5   & 0.1   \\
    Temperature ($\beta$)   & 2.0   & 2.0   & 2.0   \\
    Score Function          & MaxMC & MaxMC & MaxMC \\
    \bottomrule
\end{tabular}
\end{table}

\begin{table}[]
\centering
\caption{Hyperparameters for ACCEL across 16, 26, and 40 host networks.}
\label{tab:hyperparams_accel}
\begin{tabular}{l ccc ccc}
    \toprule
    & \multicolumn{3}{c}{\textbf{PPO ACCEL}} \\
    \cmidrule(lr){2-4}
    \textbf{Hyperparameter} & \textbf{16h} & \textbf{26h} & \textbf{40h} \\
    \midrule
    Learning Rate                   & 2e-4  & 3e-4  & 2e-4  \\
    \# Steps                        & 128   & 128   & 128   \\
    Discount Factor ($\gamma$)      & 0.99  & 0.99  & 0.99  \\
    Layer Size                      & 512   & 512   & 512   \\
    GAE Lambda ($\lambda$)          & 0.95  & 0.8   & 0.95  \\
    Clip Epsilon                    & 0.2   & 0.2   & 0.2   \\
    Entropy Coef.                   & 0.01  & 0.005 & 0.01  \\
    \# Edits                        & 10    & 5     & 5     \\
    Max Grad Norm                   & 0.5   & 0.5   & 0.5   \\
    Value Coef.                     & 0.5   & 0.5   & 0.5   \\
    Replay Prob.\ ($\rho$)          & 0.5   & 0.5   & 0.5   \\
    Staleness ($\omega$)            & 0.7   & 0.3   & 0.1   \\
    Temperature ($\beta$)           & 2.0   & 0.3   & 1.0   \\
    Score Function                  & PVL   & PVL   & PVL   \\
    Min. Fill Ratio                 & 0.3   & 0.5   & 0.5   \\
    Level Buffer Capacity           & 4000  & 10000 & 10000 \\
    \bottomrule
\end{tabular}
\end{table}

\begin{table}[]
\centering
\caption{Hyperparameters for PPO-LSTM DR across 16, 26, and 40 host networks.}
\label{tab:hyperparams_lstm_dr}
\begin{tabular}{l ccc}
    \toprule
    & \multicolumn{3}{c}{\textbf{PPO-LSTM DR}} \\
    \cmidrule(lr){2-4}
    \textbf{Hyperparameter} & \textbf{16h} & \textbf{26h} & \textbf{40h} \\
    \midrule
    Learning Rate           & 3e-4  & 3e-4  & 3e-4  \\
    \# Steps                & 64    & 32    & 64    \\
    LSTM Hidden Size        & 512   & 512   & 512   \\
    GAE Lambda ($\lambda$)  & 0.8   & 0.8   & 0.8   \\
    Clip Epsilon            & 0.1   & 0.1   & 0.1   \\
    Entropy Coef.           & 0.005 & 0.005 & 0.005 \\
    Max Grad Norm           & 1.0   & 1.0   & 0.5   \\
    Value Coef.             & 0.5   & 0.5   & 0.5   \\
    \bottomrule
\end{tabular}
\end{table}

\begin{table}[]
\centering
\caption{Hyperparameters for PPO-LSTM PLR across 16, 26, and 40 host networks.}
\label{tab:hyperparams_lstm_plr}
\begin{tabular}{l ccc}
    \toprule
    & \multicolumn{3}{c}{\textbf{PPO-LSTM PLR}} \\
    \cmidrule(lr){2-4}
    \textbf{Hyperparameter} & \textbf{16h} & \textbf{26h} & \textbf{40h} \\
    \midrule
    Learning Rate           & 2e-4  & 2e-4  & 3e-4  \\
    \# Steps                & 128   & 128   & 128   \\
    LSTM Hidden Size        & 256   & 256   & 512   \\
    GAE Lambda ($\lambda$)  & 0.8   & 0.8   & 0.8   \\
    Clip Epsilon            & 0.1   & 0.1   & 0.1   \\
    Entropy Coef.           & 0.01  & 0.01  & 0.005 \\
    Max Grad Norm           & 0.5   & 0.5   & 0.5   \\
    Value Coef.             & 0.5   & 0.5   & 0.5   \\
    Replay Prob.\ ($\rho$)  & 0.5   & 0.5   & 0.5   \\
    Staleness ($\omega$)    & 0.7   & 0.7   & 0.5   \\
    Temperature ($\beta$)   & 2.0   & 2.0   & 2.0   \\
    Score Function          & MaxMC & MaxMC & MaxMC \\
    \bottomrule
\end{tabular}
\end{table}

\begin{table}[]
\centering
\caption{Hyperparameters for PPO-LSTM PLR$^\bot$ across 16, 26, and 40 host networks.}
\label{tab:hyperparams_lstm_robust_plr}
\begin{tabular}{l ccc}
    \toprule
    & \multicolumn{3}{c}{\textbf{PPO-LSTM PLR$^\bot$}} \\
    \cmidrule(lr){2-4}
    \textbf{Hyperparameter} & \textbf{16h} & \textbf{26h} & \textbf{40h} \\
    \midrule
    Learning Rate           & 3e-4  & 2e-4  & 2e-4  \\
    \# Steps                & 128   & 128   & 128   \\
    LSTM Hidden Size        & 256   & 256   & 256   \\
    GAE Lambda ($\lambda$)  & 0.95  & 0.95  & 0.95  \\
    Clip Epsilon            & 0.2   & 0.2   & 0.2   \\
    Entropy Coef.           & 0.01  & 0.01  & 0.01  \\
    Max Grad Norm           & 0.5   & 0.5   & 0.5   \\
    Value Coef.             & 0.5   & 0.5   & 0.5   \\
    Replay Prob.\ ($\rho$)  & 0.9   & 0.9   & 0.9   \\
    Staleness ($\omega$)    & 0.3   & 0.3   & 0.3   \\
    Temperature ($\beta$)   & 1.4   & 2.0   & 2.0   \\
    Score Function          & MaxMC & MaxMC & MaxMC \\
    \bottomrule
\end{tabular}
\end{table}

\begin{table}[]
\centering
\caption{Hyperparameters for PPO-LSTM ACCEL across 16, 26, and 40 host networks.}
\label{tab:hyperparams_lstm_accel}
\begin{tabular}{l ccc}
    \toprule
    & \multicolumn{3}{c}{\textbf{PPO-LSTM ACCEL}} \\
    \cmidrule(lr){2-4}
    \textbf{Hyperparameter} & \textbf{16h} & \textbf{26h} & \textbf{40h} \\
    \midrule
    Learning Rate           & 3e-4  & 3e-4  & 3e-4  \\
    \# Steps                & 128   & 128   & 128   \\
    LSTM Hidden Size        & 512   & 512   & 512   \\
    GAE Lambda ($\lambda$)  & 0.8   & 0.8   & 0.8   \\
    Clip Epsilon            & 0.3   & 0.1   & 0.1   \\
    Entropy Coef.           & 0.01  & 0.005 & 0.005 \\
    Max Grad Norm           & 0.5   & 0.5   & 0.5   \\
    Value Coef.             & 0.5   & 0.5   & 0.5   \\
    Replay Prob.\ ($\rho$)  & 0.5   & 0.5   & 0.5   \\
    Staleness ($\omega$)    & 0.7   & 0.1   & 0.5   \\
    Temperature ($\beta$)   & 0.1   & 0.3   & 0.1   \\
    Score Function          & MaxMC & MaxMC & PVL   \\
    \# Edits                & 10    & 10    & 10    \\
    Min.\ Fill Ratio        & 0.3   & 0.5   & 0.5   \\
    \bottomrule
\end{tabular}
\end{table}

\section{Active Host Distributions}\label{app:active_host_distributions}

\begin{figure*}[ht]
    \centering
    \begin{subfigure}[t]{0.45\linewidth}
        \centering
        \includegraphics[width=\linewidth]{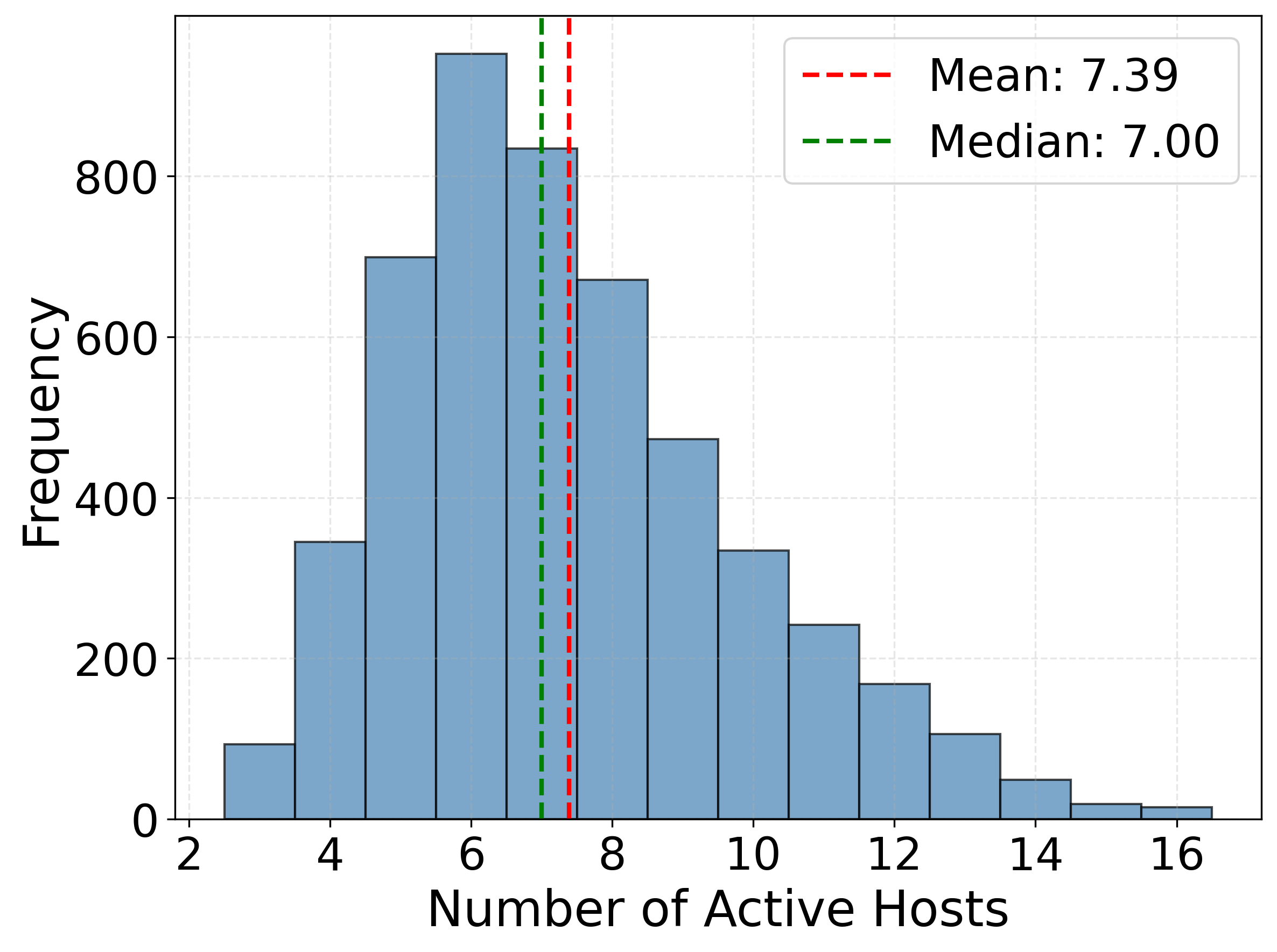}
        \caption{$t_d=0.06$}
        \label{fig:network_ahd_16_hosts_td_006}
    \end{subfigure}
    \hfill
    \begin{subfigure}[t]{0.45\linewidth}
        \centering
        \includegraphics[width=\linewidth]{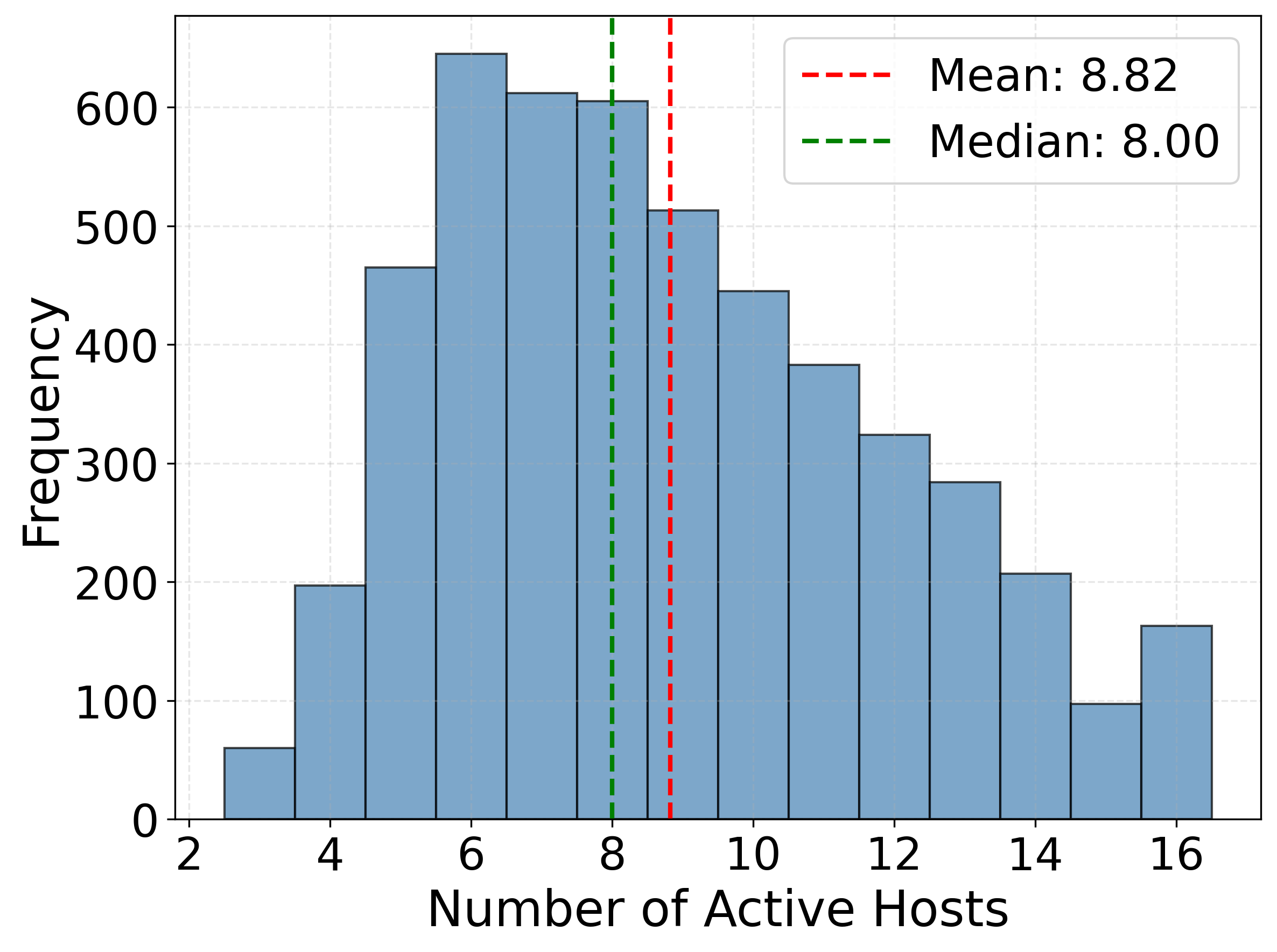}
        \caption{$t_d=0.115$}
        \label{fig:network_ahd_16_hosts_td_0115}
    \end{subfigure}
    \hfill
    \begin{subfigure}[t]{0.45\linewidth}
        \centering
        \includegraphics[width=\linewidth]{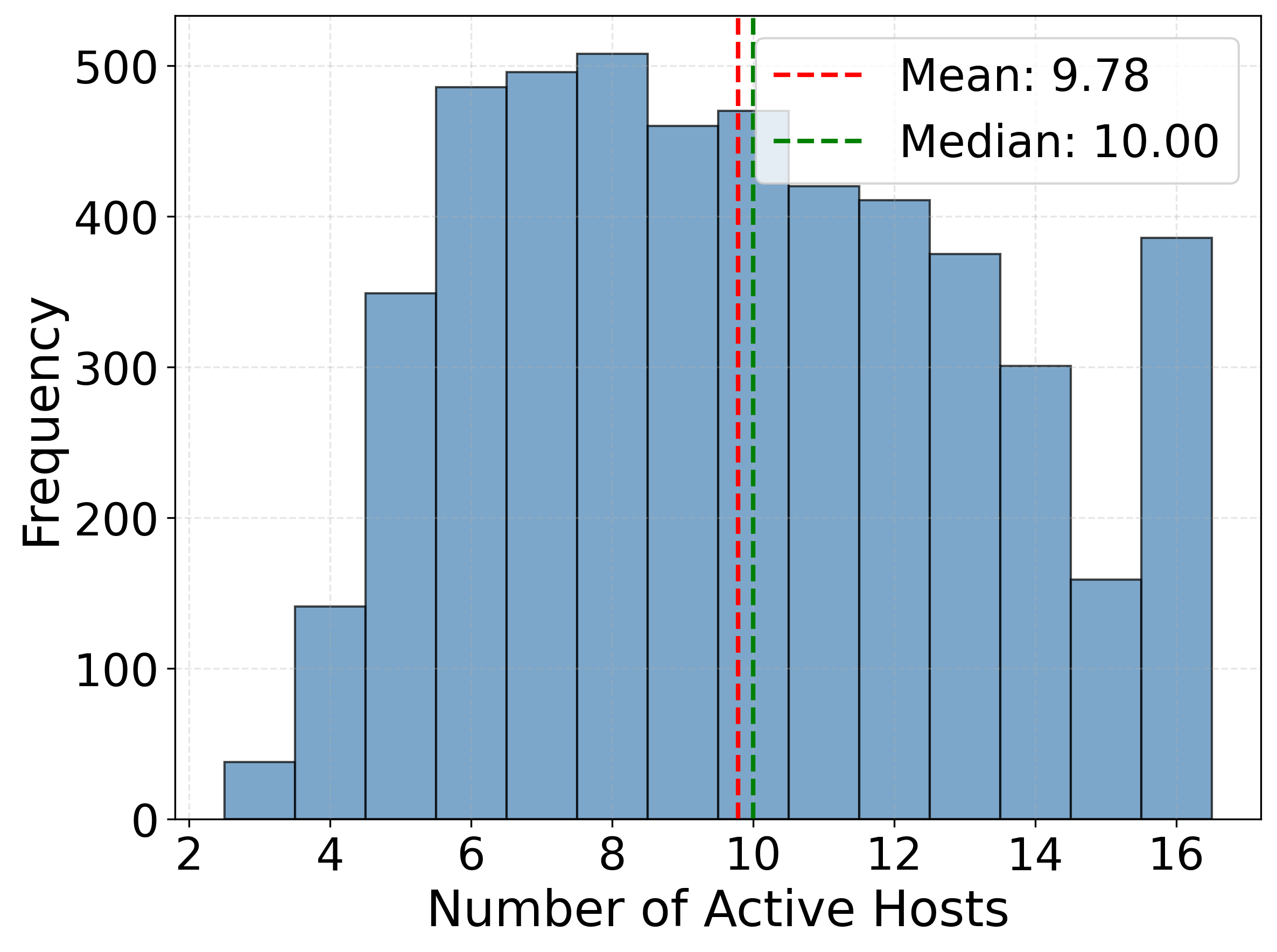}
        \caption{$t_d=0.15$}
        \label{fig:network_ahd_16_hosts_td_015}
    \end{subfigure}
    \hfill
    \begin{subfigure}[t]{0.45\linewidth}
        \centering
        \includegraphics[width=\linewidth]{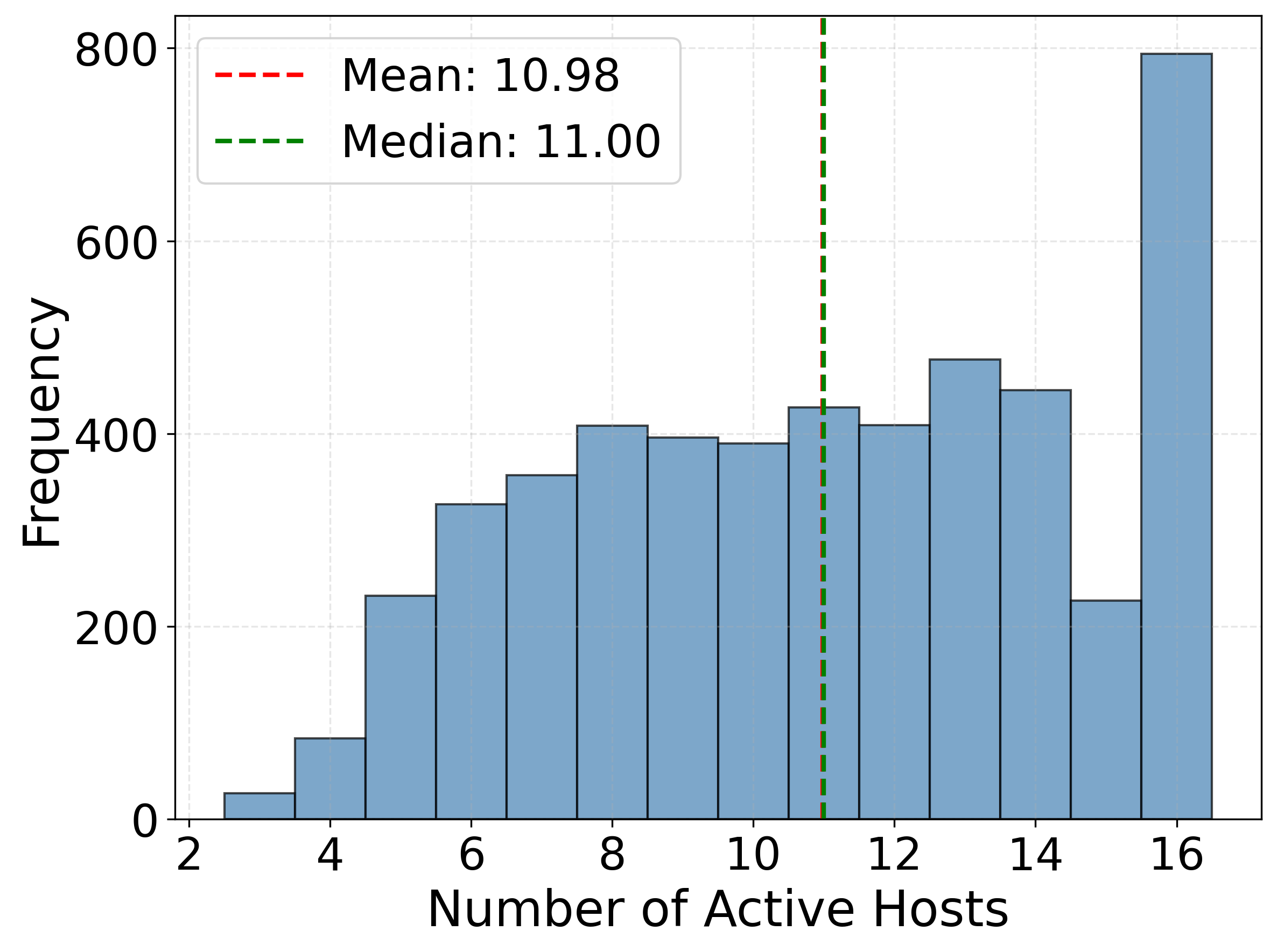}
        \caption{$t_d=0.195$}
        \label{fig:network_ahd_16_hosts_td_0195}
    \end{subfigure}
    \hfill
    \begin{subfigure}[t]{0.45\linewidth}
        \centering
        \includegraphics[width=\linewidth]{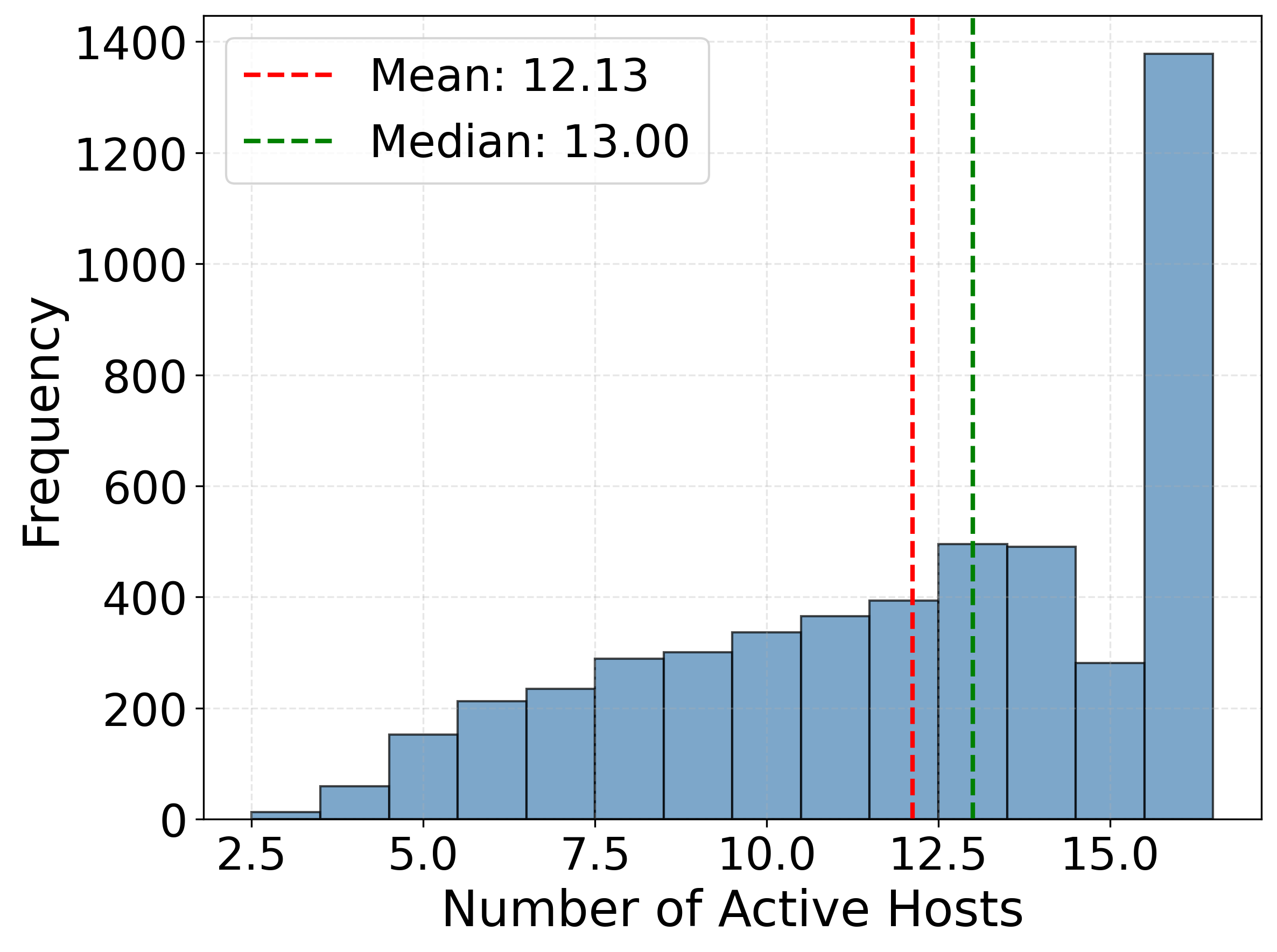}
        \caption{$t_d=0.24$}
        \label{fig:network_ahd_16_hosts_td_024}
    \end{subfigure}
    \hfill
    
    \caption{Distributions of active hosts within 16 host networks and how the topology density ($t_d$) parameter influences them.}
    \label{fig:network_ahd_16_hosts}
\end{figure*}

\begin{figure*}[ht]
    \centering
    \begin{subfigure}[t]{0.45\linewidth}
        \centering
        \includegraphics[width=\linewidth]{figures/network_ahd_26_hosts/network_ahd_26_hosts_td_0.04.png}
        \caption{$t_d=0.04$}
        \label{fig:network_ahd_26_hosts_td_004}
    \end{subfigure}
    \hfill
    \begin{subfigure}[t]{0.45\linewidth}
        \centering
        \includegraphics[width=\linewidth]{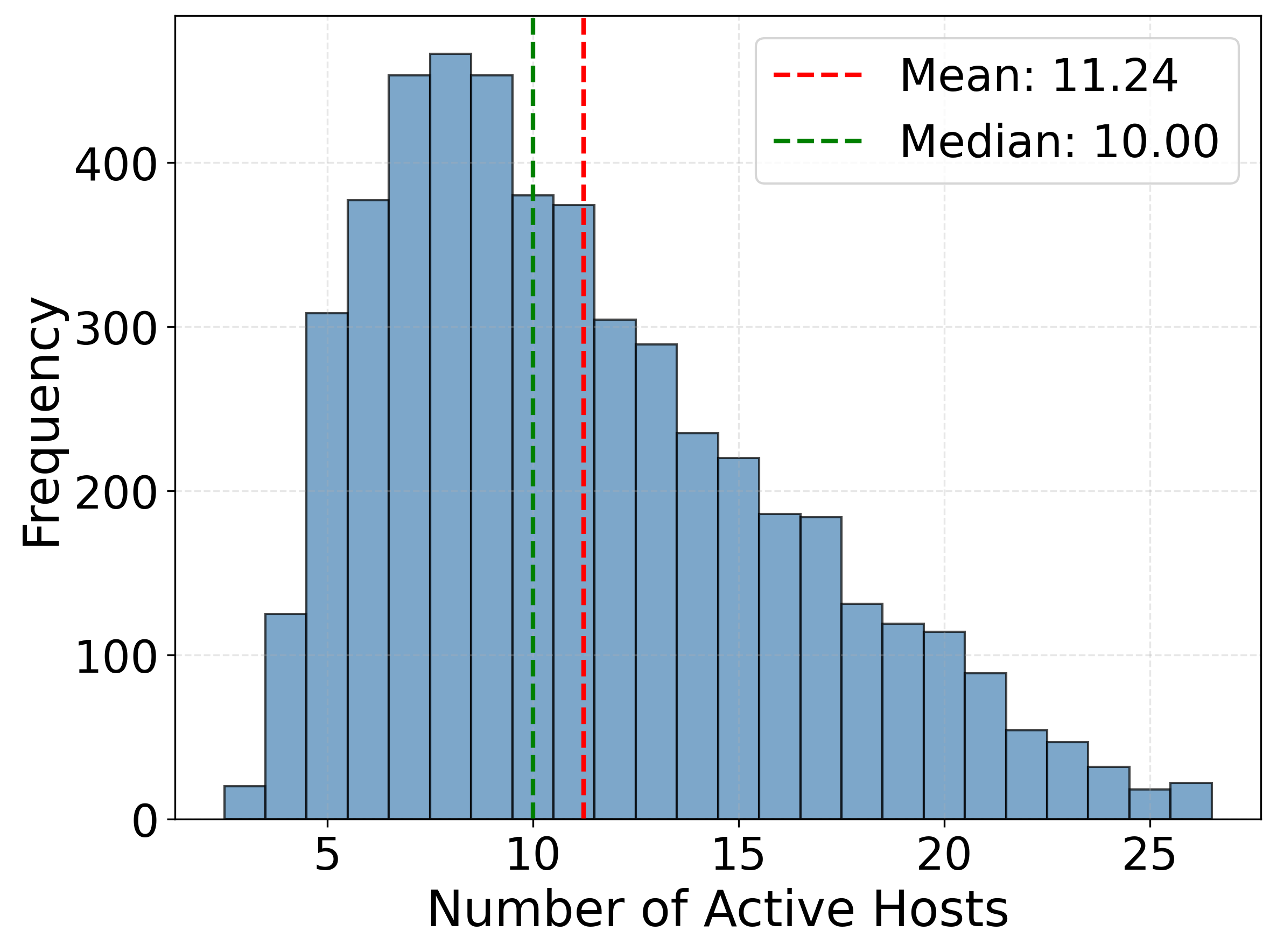}
        \caption{$t_d=0.08$}
        \label{fig:network_ahd_26_hosts_td_008}
    \end{subfigure}
    \hfill
    \begin{subfigure}[t]{0.45\linewidth}
        \centering
        \includegraphics[width=\linewidth]{figures/network_ahd_26_hosts/network_ahd_26_hosts_td_0.12.png}
        \caption{$t_d=0.12$}
        \label{fig:network_ahd_26_hosts_td_012}
    \end{subfigure}
    \hfill
    \begin{subfigure}[t]{0.45\linewidth}
        \centering
        \includegraphics[width=\linewidth]{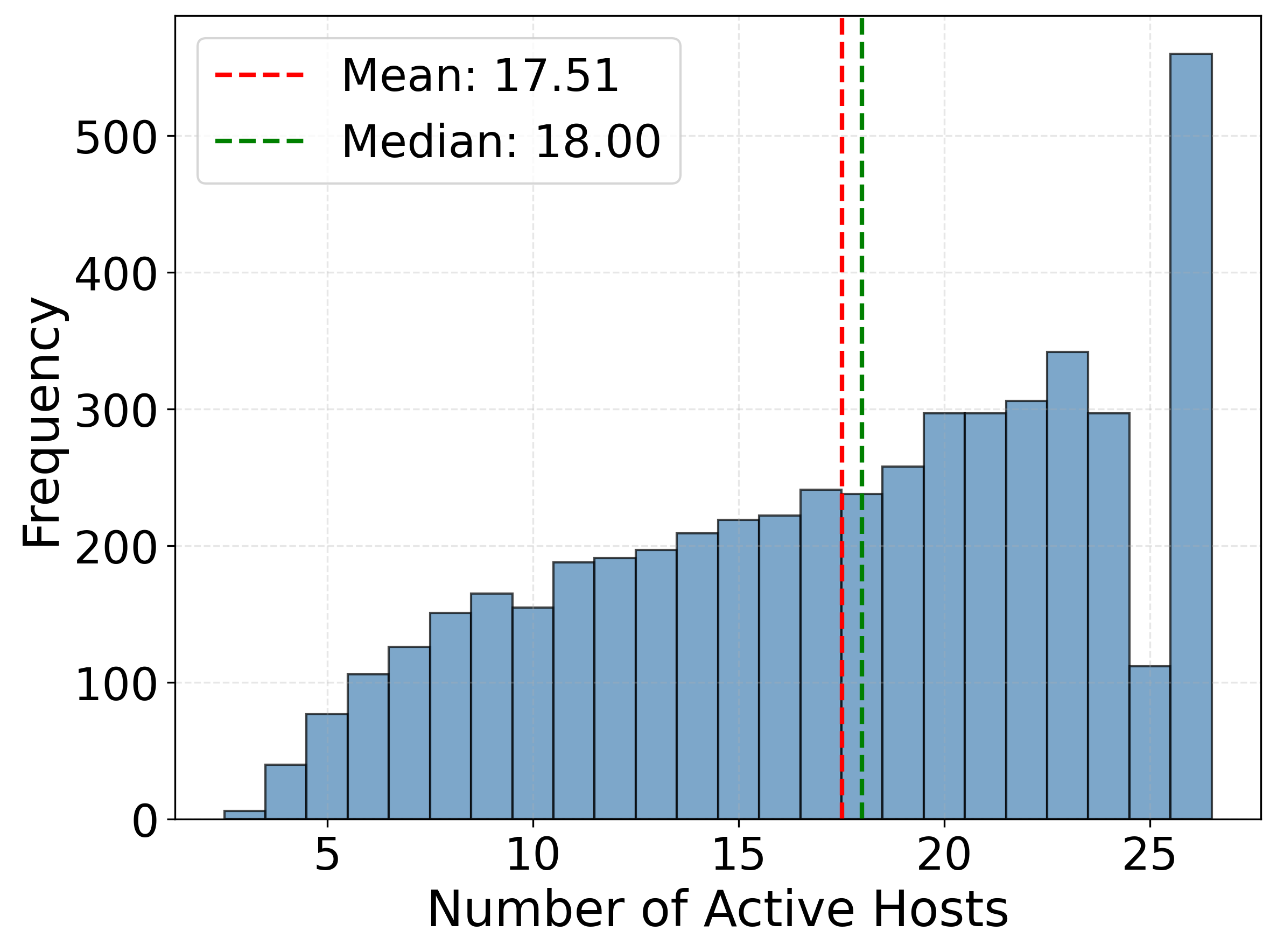}
        \caption{$t_d=0.16$}
        \label{fig:network_ahd_26_hosts_td_016}
    \end{subfigure}
    \hfill
    \begin{subfigure}[t]{0.45\linewidth}
        \centering
        \includegraphics[width=\linewidth]{figures/network_ahd_26_hosts/network_ahd_26_hosts_td_0.2.png}
        \caption{$t_d=0.2$}
        \label{fig:network_ahd_26_hosts_td_02}
    \end{subfigure}
    \hfill
    
    \caption{Distributions of active hosts within 26 host networks and how the topology density ($t_d$) parameter influences them.}
    \label{fig:network_ahd_26_hosts}
\end{figure*}

\begin{figure*}[ht]
    \centering
    \begin{subfigure}[t]{0.45\linewidth}
        \centering
        \includegraphics[width=\linewidth]{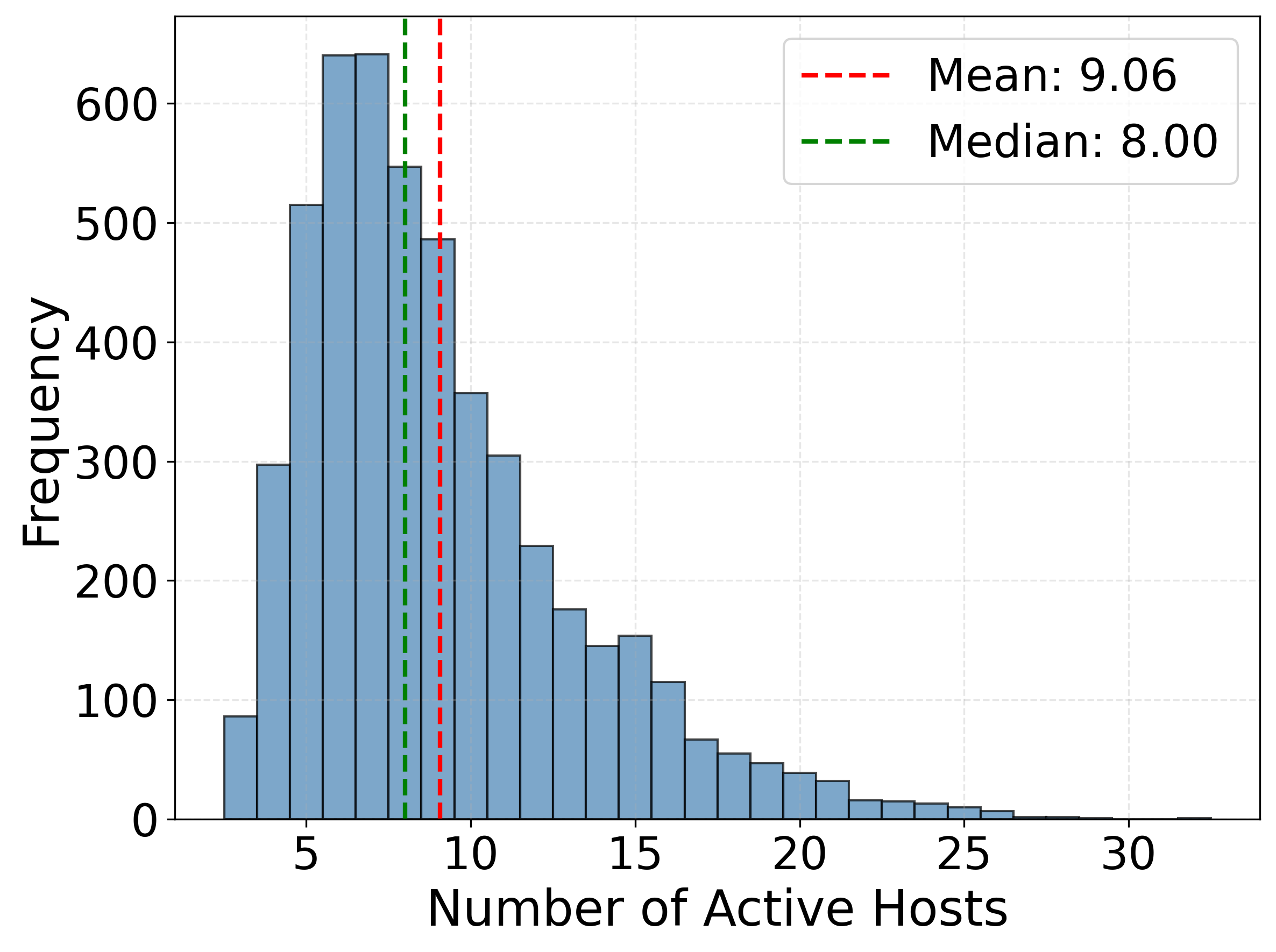}
        \caption{$t_d=0.03$}
        \label{fig:network_ahd_40_hosts_td_003}
    \end{subfigure}
    \hfill
    \begin{subfigure}[t]{0.45\linewidth}
        \centering
        \includegraphics[width=\linewidth]{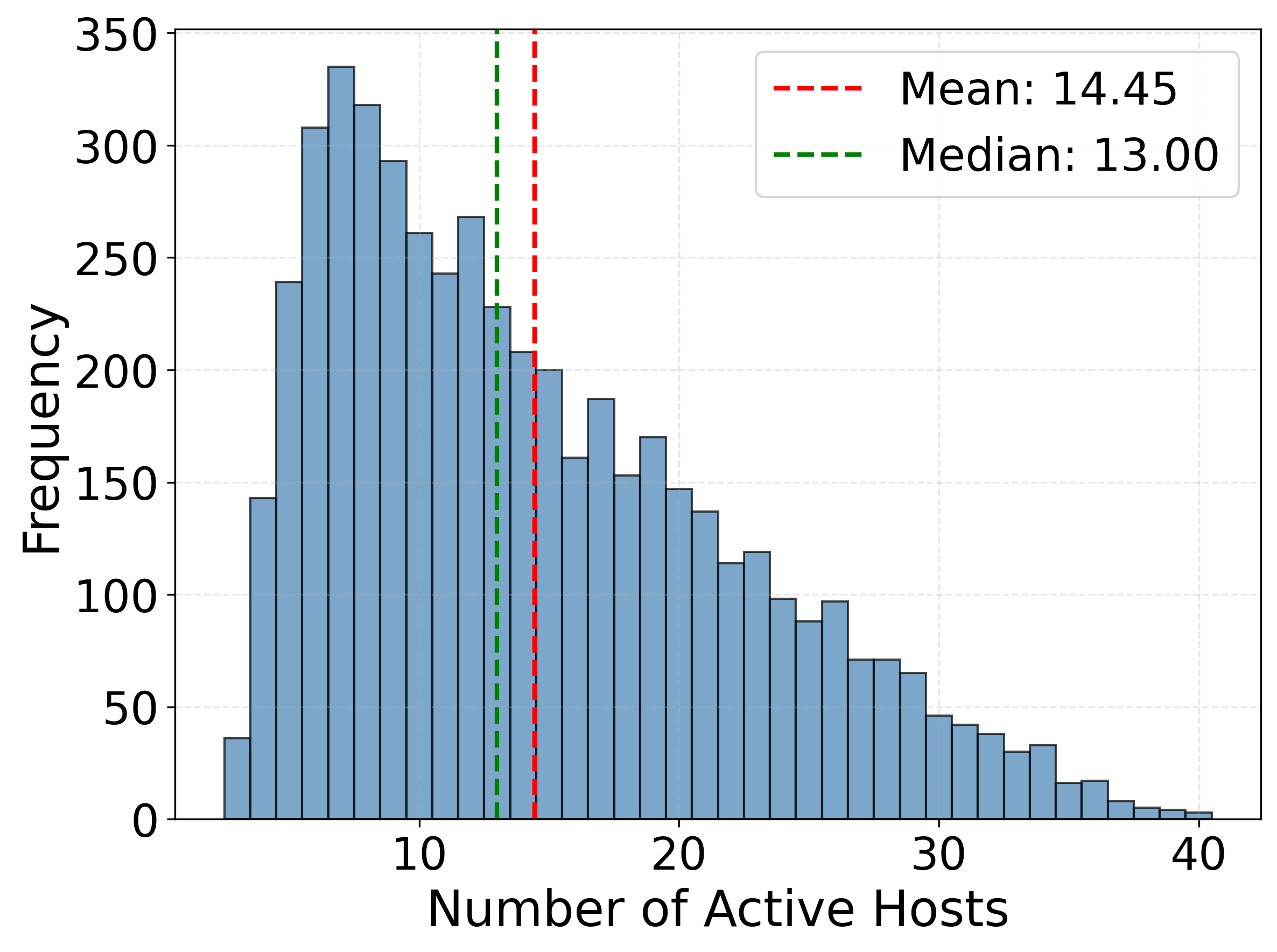}
        \caption{$t_d=0.06$}
        \label{fig:network_ahd_40_hosts_td_006}
    \end{subfigure}
    \hfill
    \begin{subfigure}[t]{0.45\linewidth}
        \centering
        \includegraphics[width=\linewidth]{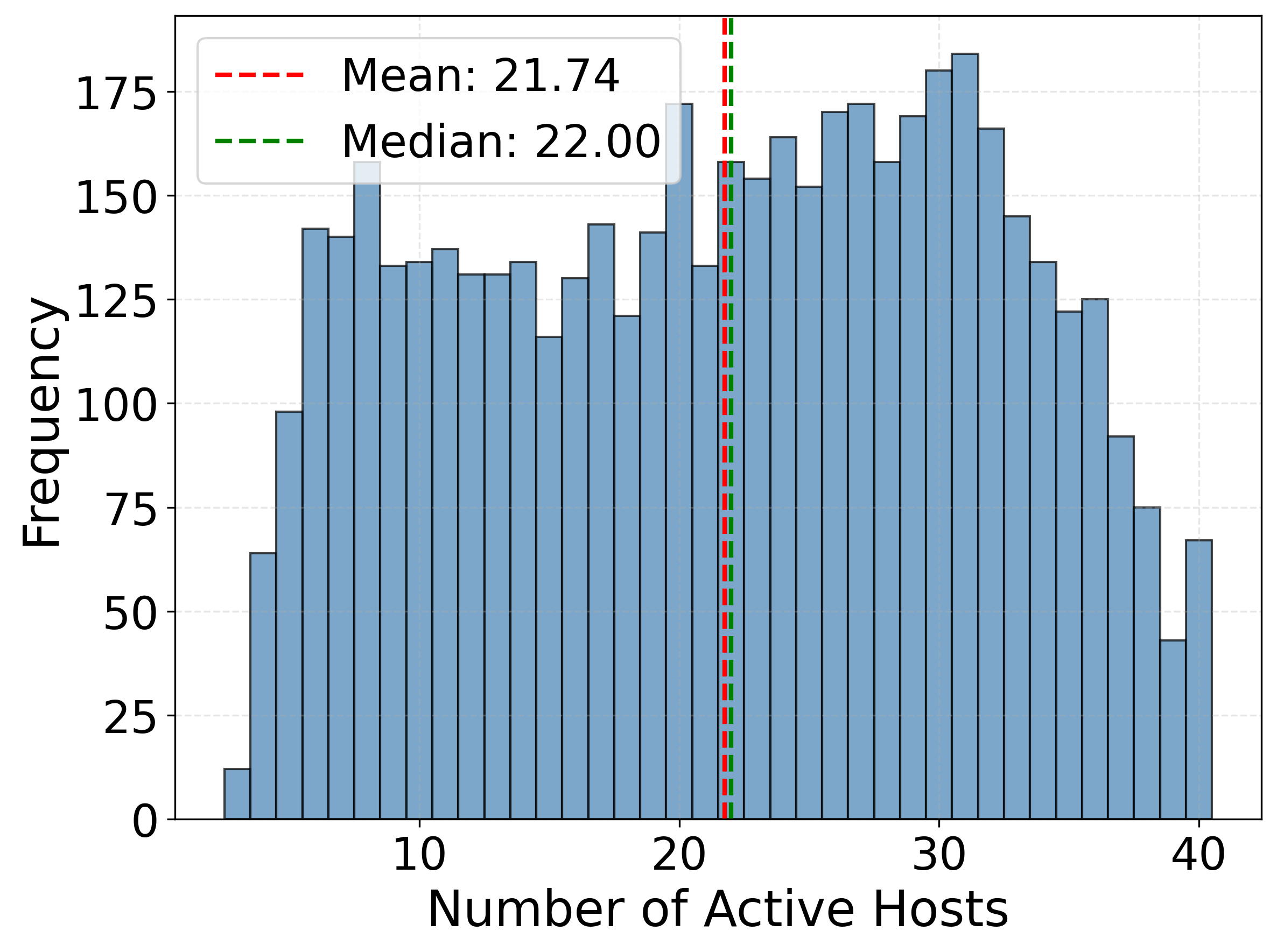}
        \caption{$t_d=0.09$}
        \label{fig:network_ahd_40_hosts_td_009}
    \end{subfigure}
    \hfill
    \begin{subfigure}[t]{0.45\linewidth}
        \centering
        \includegraphics[width=\linewidth]{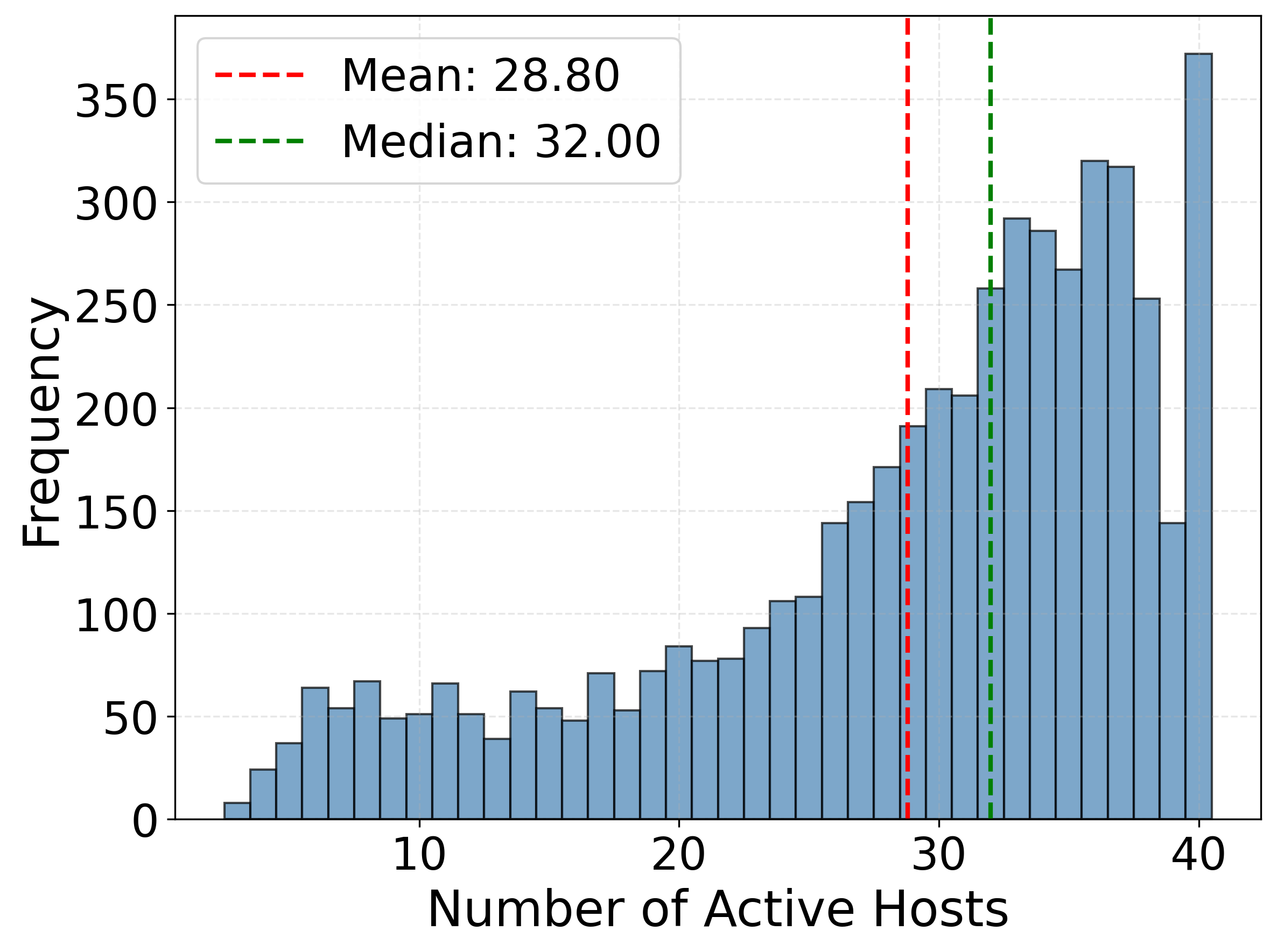}
        \caption{$t_d=0.12$}
        \label{fig:network_ahd_40_hosts_td_012}
    \end{subfigure}
    \hfill
    \begin{subfigure}[t]{0.45\linewidth}
        \centering
        \includegraphics[width=\linewidth]{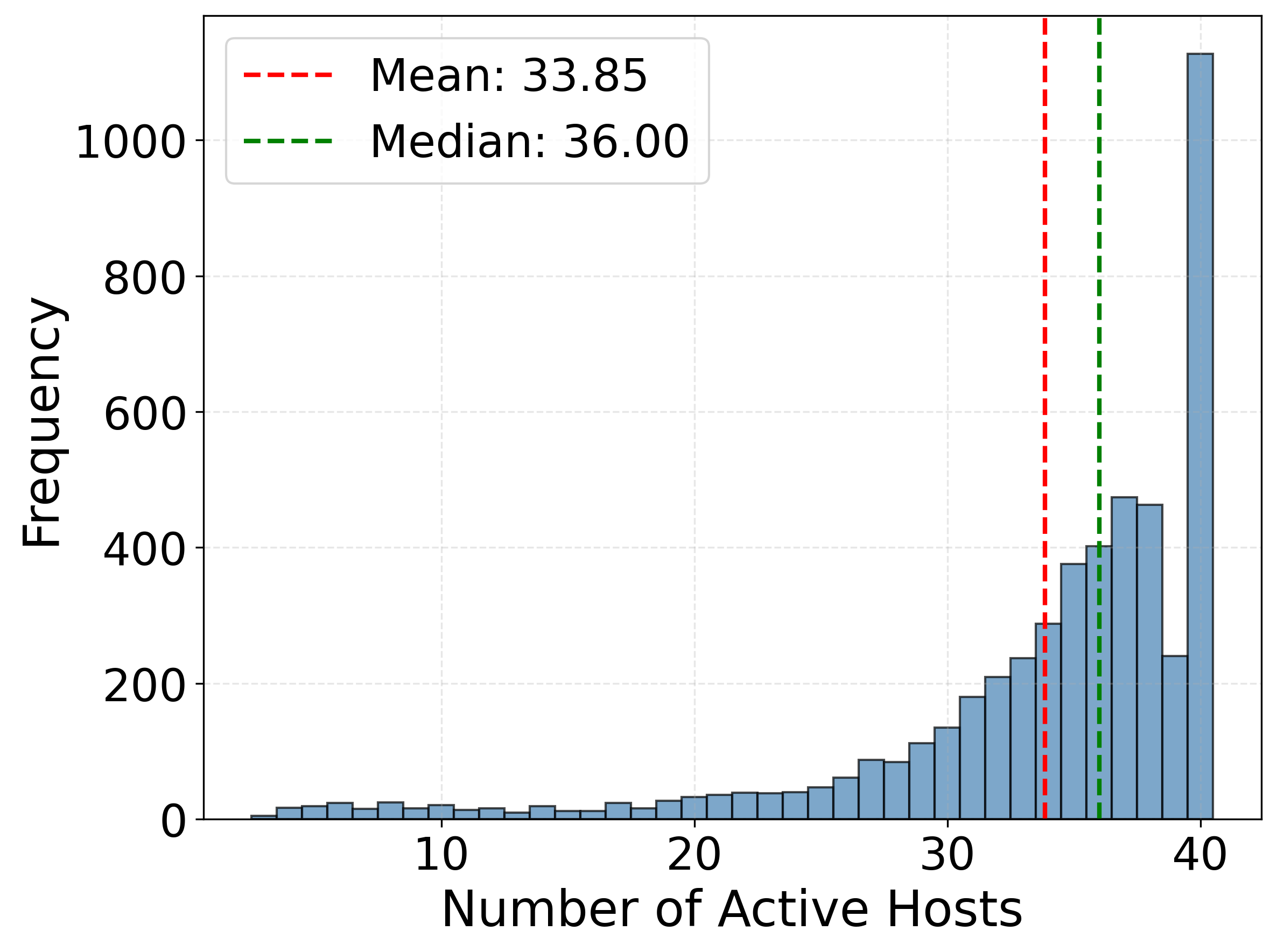}
        \caption{$t_d=0.15$}
        \label{fig:network_ahd_40_hosts_td_015}
    \end{subfigure}
    \hfill
    
    \caption{Distributions of active hosts for 40 host networks and how the topology density ($t_d$) parameter influences them.}
    \label{fig:network_ahd_40_hosts}
\end{figure*}

Here we present the full set of distributions for the $t_d$-values established in Table \ref{tab:env_config}. As we've selected to use five values for each number of total hosts in the network, we present the resulting distributions in Figures \ref{fig:network_ahd_16_hosts}, \ref{fig:network_ahd_26_hosts} and \ref{fig:network_ahd_40_hosts}. These clearly showcase the shift in active hosts given $t_d$.

\section{Extended Results}\label{app:extended_results}

In this Section we highlight some extended results for the experiments conducted in Section \ref{sec:exp_zspt}, which showcases the aggregated evaluation performance given different training $t_d$-values. Figures~\ref{fig:topo_density_eval_curves_16h_complete}, \ref{fig:topo_density_eval_curves_26h_complete} \ref{fig:topo_density_eval_curves_40h_complete} display more detailed results, where we plot for every algorithm and $t_d$-train value the performance against the evaluation $t_d$-values.

\begin{figure*}
    \centering
    \includegraphics[width=\linewidth]{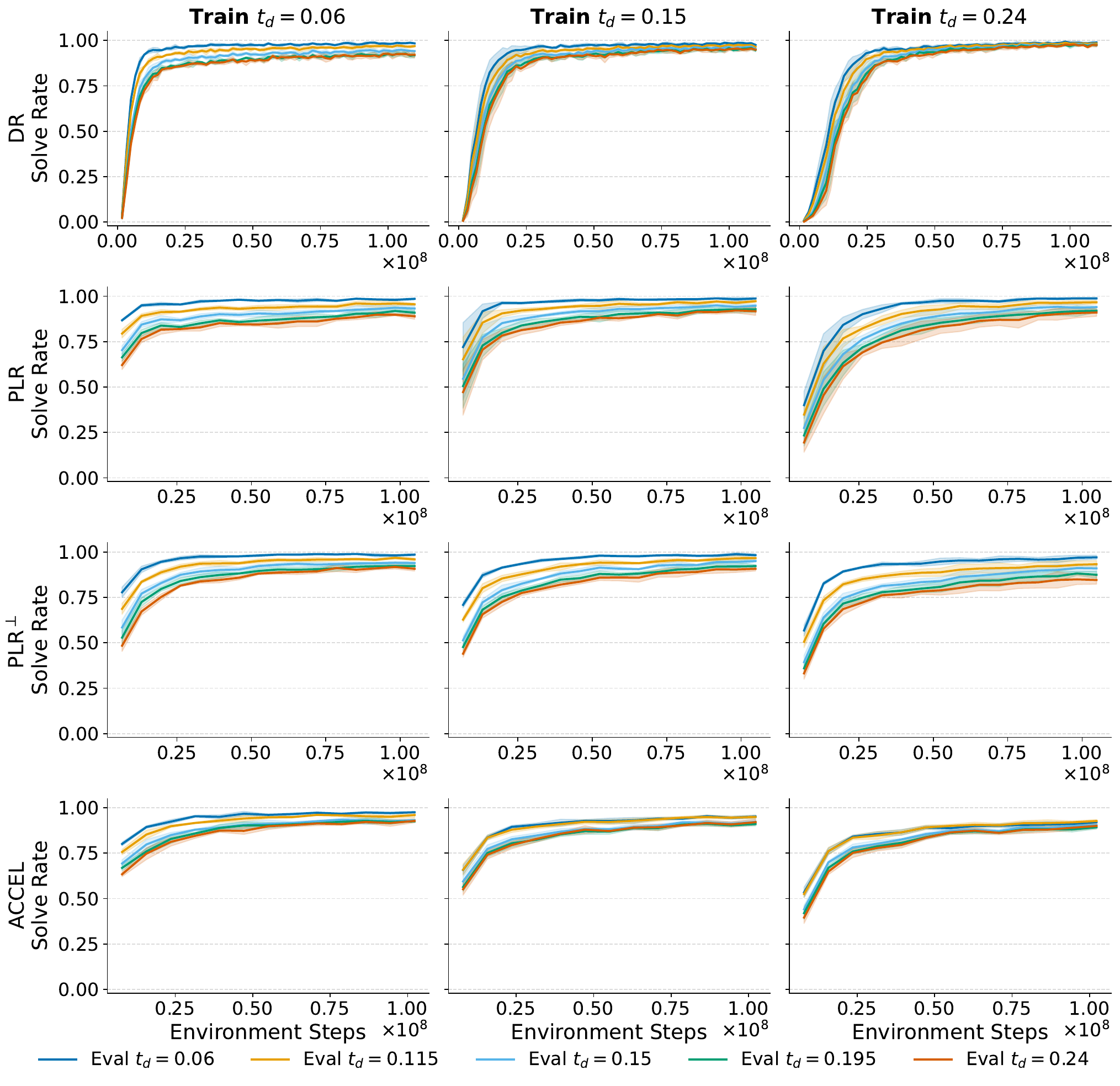}
    \caption{Evaluation curves for 16 host networks. Every column represents the network's $t_d$ the policy was trained on. The plots show the evaluation performance against all other selected $t_d$ for the given network size, as displayed in Table \ref{tab:env_config}.}
    \label{fig:topo_density_eval_curves_16h_complete}
\end{figure*}

\begin{figure*}
    \centering
    \includegraphics[width=\linewidth]{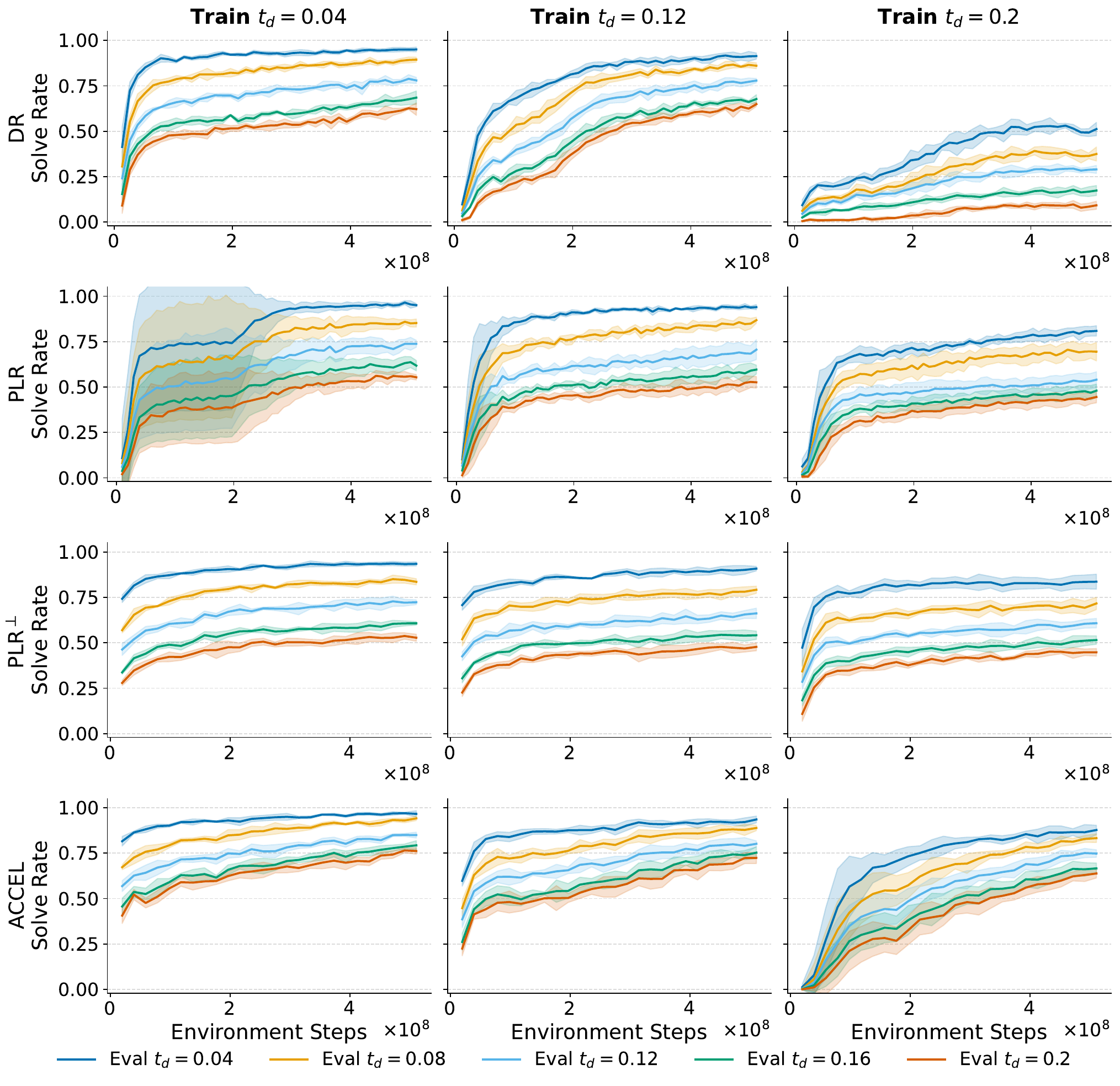}
    \caption{Evaluation curves for 26 host networks. Every column represents the network's $t_d$ the policy was trained on. The plots show the evaluation performance against all other selected $t_d$ for the given network size, as displayed in Table \ref{tab:env_config}.}
    \label{fig:topo_density_eval_curves_26h_complete}
\end{figure*}

\begin{figure*}
    \centering
    \includegraphics[width=\linewidth]{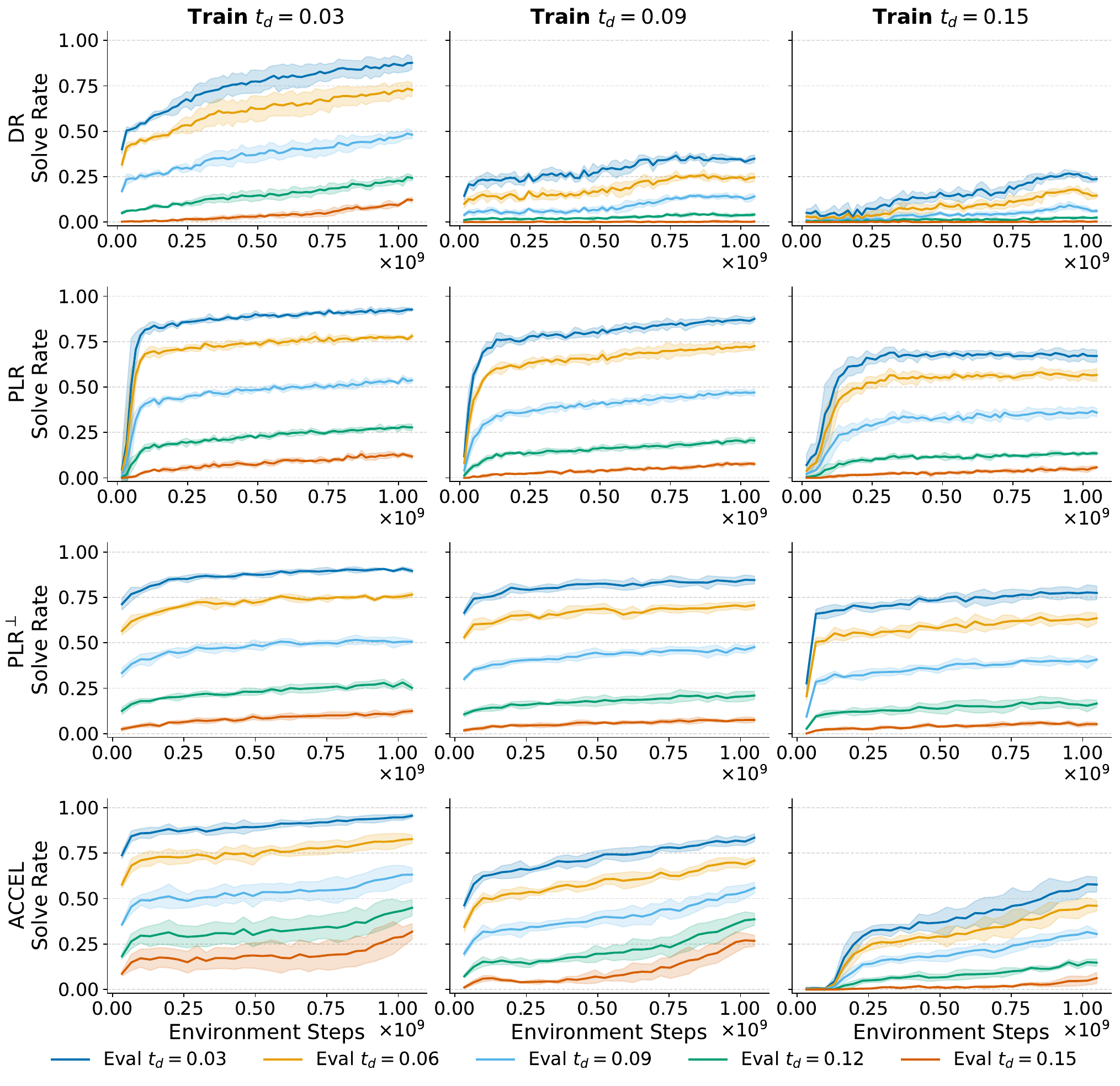}
    \caption{Evaluation curves for 40 host networks. Every column represents the network's $t_d$ the policy was trained on. The plots show the evaluation performance against all other selected $t_d$ for the given network size, as displayed in Table \ref{tab:env_config}.}
    \label{fig:topo_density_eval_curves_40h_complete}
\end{figure*}

\begin{figure*}
    \centering
    \includegraphics[width=\linewidth]{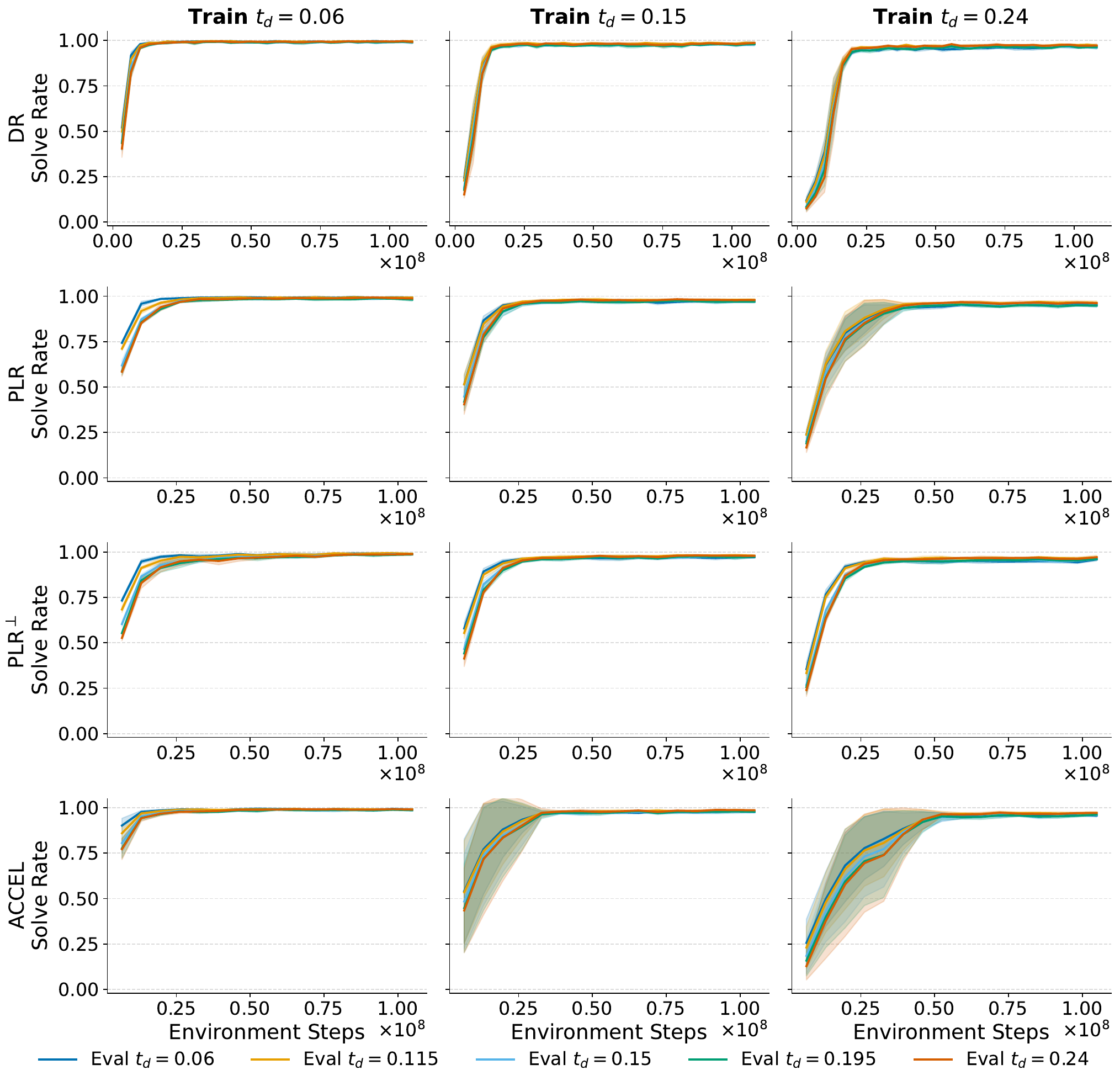}
    \caption{Evaluation curves for 16 host networks (PPO-LSTM). Every column represents the network's $t_d$ the policy was trained on. The plots show the evaluation performance against all other selected $t_d$ for the given network size, as displayed in Table \ref{tab:env_config}.}
    \label{fig:topo_density_eval_curves_16h_complete}
\end{figure*}

\begin{figure*}
    \centering
    \includegraphics[width=\linewidth]{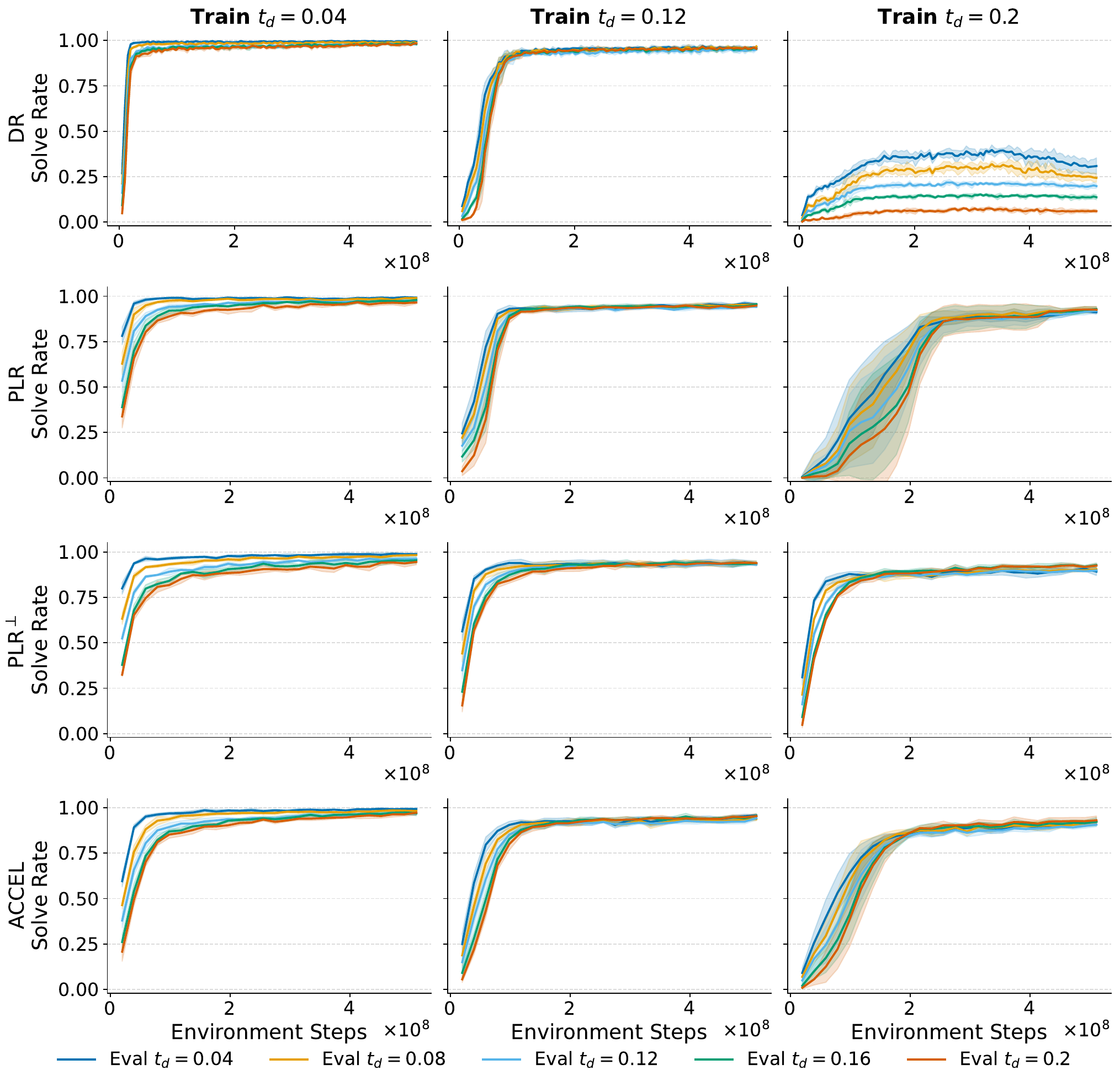}
    \caption{Evaluation curves for 26 host networks (PPO-LSTM). Every column represents the network's $t_d$ the policy was trained on. The plots show the evaluation performance against all other selected $t_d$ for the given network size, as displayed in Table \ref{tab:env_config}.}
    \label{fig:topo_density_eval_curves_26h_complete}
\end{figure*}

\begin{figure*}
    \centering
    \includegraphics[width=\linewidth]{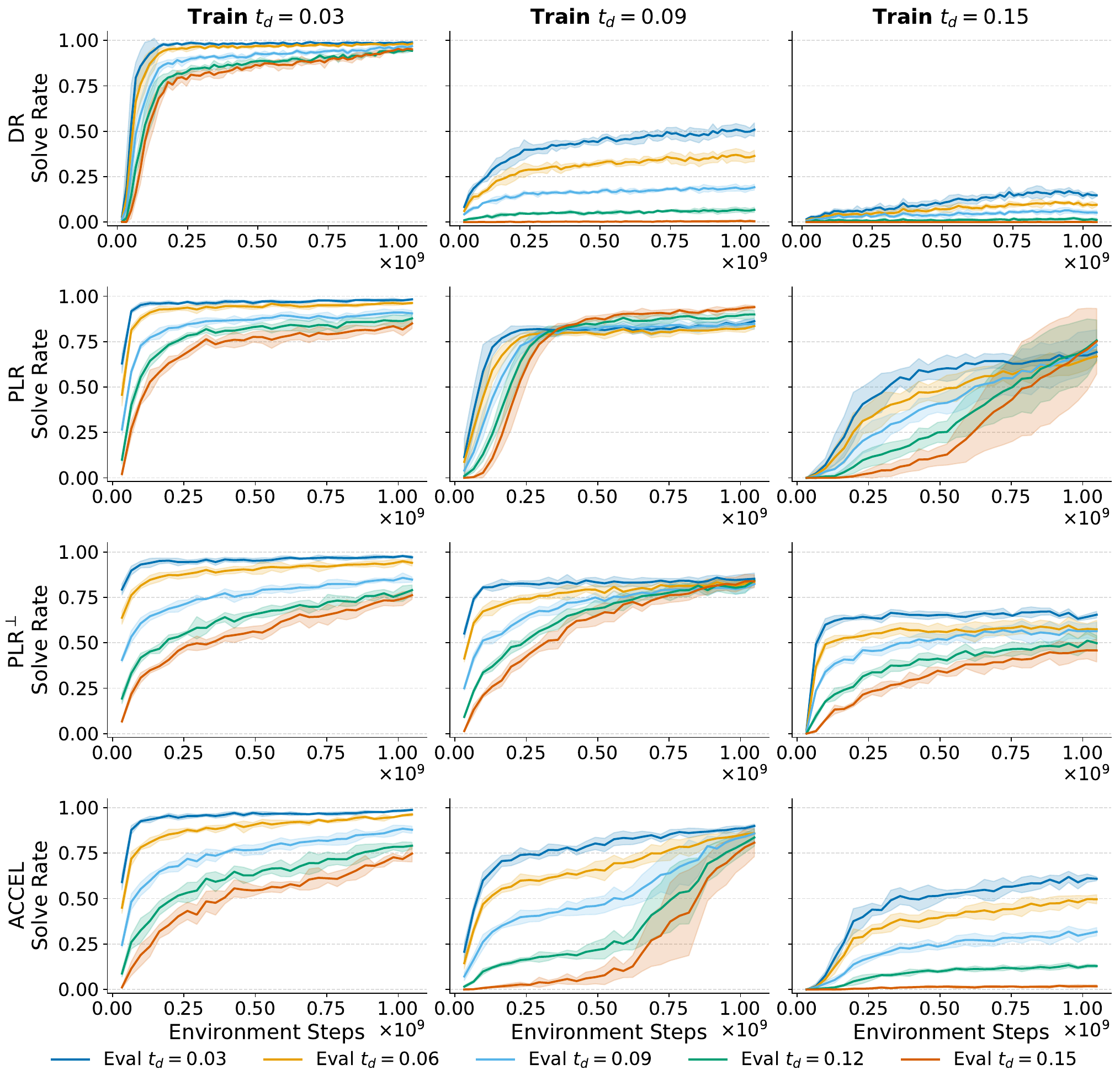}
    \caption{Evaluation curves for 40 host networks (PPO-LSTM). Every column represents the network's $t_d$ the policy was trained on. The plots show the evaluation performance against all other selected $t_d$ for the given network size, as displayed in Table \ref{tab:env_config}.}
    \label{fig:topo_density_eval_curves_40h_complete}
\end{figure*}

\section{Computational resources}\label{app:computation_resources}

In this Section we present the computational resources used to produce the results presented in the paper. Table \ref{tab:hardware_cluster} presents the hardware configurations of the compute clusters we had access to. Table \ref{tab:hardware_workstations} presents the hardware configurations of the compute workstations \& servers. Table \ref{tab:compute} discloses the total wall-clock time necessary to reproduce the results of the experiments. Table~\ref{tab:sweep-compute} reports the total wall-clock time and the number of configurations evaluated for each hyperparameter sweep.

\begin{table}[t]
  \centering
  \caption{Composition and hardware configuration of the cluster compute nodes partially used for experiments.}
  \label{tab:hardware_cluster}
  \small
  \begin{tabular}{l l l l l l}
    \toprule
    \textbf{Cluster} & \textbf{GPU \#} & \textbf{CPU} & \textbf{GPU} & \textbf{VRAM} & \textbf{RAM} \\
    \midrule
    Cluster 1 & Node 1 & 64 core CPU   & 4xNVIDIA H100      & 372\,GB & 251\,GB \\
              & Node 2 & 64 core CPU   & 4xNVIDIA A100      & 320\,GB & 377\,GB \\
    \midrule
    Cluster 2 & Node 1 & 2xAMD EPYC 9275F (24C) & 2xNVIDIA H200 & 280\,GB & 754\,GB \\
    \bottomrule
  \end{tabular}
\end{table}

\begin{table}[t]
  \centering
  \caption{Hardware configuration of the compute machines used for experiments.}
  \label{tab:hardware_workstations}
  \small
  \begin{tabular}{l l l l l}
    \toprule
    \textbf{Machine} & \textbf{CPU} & \textbf{GPU} & \textbf{VRAM} & \textbf{RAM} \\
    \midrule
    6 Workstations & AMD Ryzen 9 9950x (16C)     & NVIDIA RTX 5090        & 32\,GB & 64\,GB \\
    Server 1 & Intel(R) Xeon(R) Gold 6230R (26C)   & NVIDIA RTX A4000   & 16\,GB & 128\,GB \\
    Server 2 & AMD EPYC 7413 (8VCPU Cores)   & NVIDIA L40S   & 48\,GB & 96\,GB \\
    \bottomrule
  \end{tabular}
\end{table}

\begin{table}[t]
\centering
\caption{Compute used to produce the experimental results. ``GPU(s)'' lists every distinct device that contributed; ``Runtime'' is the total wall-clock time summed over all seeds reported.}
\label{tab:compute}
\small
\begin{tabular}{l l r}
\toprule
\textbf{Experiment} & \textbf{GPU(s)} & \textbf{Runtime (all seeds)} \\
\midrule
Baseline PPO -- 16 hosts & RTX 5090 & 57.0\,min \\
Baseline PQN -- 16 hosts & RTX 5090 & 1.2\,h \\
Baseline PPO -- 26 hosts & RTX 5090 & 3.2\,h \\
Baseline PQN -- 26 hosts & RTX 5090 & 3.2\,h \\
Baseline PPO -- 40 hosts & RTX 5090 & 7.7\,h \\
Baseline PQN -- 40 hosts & RTX 5090 & 7.5\,h \\
Topo-density UED (PPO) -- DR -- 16 hosts & H100 NVL & 2.6\,h \\
Topo-density UED (PPO) -- PLR -- 16 hosts & H100 NVL & 1.1\,h \\
Topo-density UED (PPO) -- PLR$^\bot$ -- 16 hosts & H100 NVL & 1.3\,h \\
Topo-density UED (PPO) -- ACCEL -- 16 hosts & RTX 5090, RTX A4000 & 2.9\,h \\
Topo-density UED (PPO) -- DR -- 26 hosts & H100 NVL & 7.6\,h \\
Topo-density UED (PPO) -- PLR -- 26 hosts & H100 NVL & 6.8\,h \\
Topo-density UED (PPO) -- PLR$^\bot$ -- 26 hosts & H100 NVL & 5.0\,h \\
Topo-density UED (PPO) -- ACCEL -- 26 hosts & RTX 5090 & 5.2\,h \\
Topo-density UED (PPO) -- DR -- 40 hosts & RTX 5090 & 41.2\,h \\
Topo-density UED (PPO) -- PLR -- 40 hosts & RTX 5090 & 36.7\,h \\
Topo-density UED (PPO) -- PLR$^\bot$ -- 40 hosts & A100-PCIE-40GB, RTX 5090 & 39.5\,h \\
Topo-density UED (PPO) -- ACCEL -- 40 hosts & L40S, RTX 5090 & 18.0\,h \\
\midrule Topo-density UED (LSTM) -- DR -- 16 hosts & RTX A4000 & 11.3\,h \\
Topo-density UED (LSTM) -- PLR -- 16 hosts & RTX A4000 & 7.2\,h \\
Topo-density UED (LSTM) -- PLR$^\bot$ -- 16 hosts & RTX 5090 & 3.9\,h \\
Topo-density UED (LSTM) -- ACCEL -- 16 hosts & RTX 5090 & 4.2\,h \\
Topo-density UED (LSTM) -- DR -- 26 hosts & RTX 5090 & 21.3\,h \\
Topo-density UED (LSTM) -- PLR -- 26 hosts & RTX 5090, RTX 6000 Ada Generation & 16.5\,h \\
Topo-density UED (LSTM) -- PLR$^\bot$ -- 26 hosts & RTX 5090, RTX 6000 Ada Generation & 16.3\,h \\
Topo-density UED (LSTM) -- ACCEL -- 26 hosts & H100 NVL, RTX 5090 & 20.2\,h \\
Topo-density UED (LSTM) -- DR -- 40 hosts & RTX 5090 & 40.6\,h \\
Topo-density UED (LSTM) -- PLR -- 40 hosts & RTX 5090, RTX 6000 Ada Generation & 38.8\,h \\
Topo-density UED (LSTM) -- PLR$^\bot$ -- 40 hosts & H100 NVL, H200 NVL, RTX 5090 & 41.4\,h \\
Topo-density UED (LSTM) -- ACCEL -- 40 hosts & RTX 5090 & 38.3\,h \\
\midrule \textbf{Total} & (437 runs) & \textbf{18.8\,d (451\,h)} \\
\bottomrule
\end{tabular}
\end{table}

\begin{table}[t]
\centering
\caption{Compute used for the hyperparameter sweeps per algorithm.}
\label{tab:sweep-compute}
\small
\begin{tabular}{l l r r}
\toprule
\textbf{Algorithm} & \textbf{GPU(s)} & \textbf{\# Runs} & \textbf{Runtime (all runs)} \\
\midrule
PQN & RTX 5090, RTX 6000 Ada Generation, RTX A4000 & 383 & 149\,h \\
ACCEL & A100-SXM4-80GB, H200 NVL, RTX 5090, RTX A4000 & 275 & 82.3\,h \\
Robust PLR & A100-SXM4-80GB, H100 NVL, RTX 5090 & 471 & 250\,h \\
PLR & A100-SXM4-80GB, H100 NVL, RTX 5090 & 476 & 305\,h \\
PPO DR & H100 NVL, RTX 5090 & 331 & 129\,h \\
\midrule PPO-LSTM DR & RTX 5090, RTX A4000 & 38 & 35.3\,h \\
PPO-LSTM PLR & RTX 5090 & 58 & 55.3\,h \\
PPO-LSTM Robust PLR & RTX 5090 & 48 & 46.0\,h \\
PPO-LSTM ACCEL & \makecell{H100 NVL, H200 NVL, RTX 5090, \\ RTX 6000 Ada Generation, RTX A4000} & 258 & 238\,h \\
\midrule \textbf{Total} &  & \textbf{2338} & \textbf{1289\,h} \\
\bottomrule
\end{tabular}
\end{table}

%% file: bibliography.bib
@misc{standen_cyborg_2021,
	title = {{CybORG}: {A} {Gym} for the {Development} of {Autonomous} {Cyber} {Agents}},
	shorttitle = {{CybORG}},
	urldate = {2023-11-17},
	publisher = {arXiv},
	author = {Standen, Maxwell and Lucas, Martin and Bowman, David and Richer, Toby J. and Kim, Junae and Marriott, Damian},
	month = aug,
	year = {2021},
	note = {arXiv:2108.09118 [cs]},
}

@misc{janisch_nasimemu_2023,
	title = {{NASimEmu}: {Network} {Attack} {Simulator} \& {Emulator} for {Training} {Agents} {Generalizing} to {Novel} {Scenarios}},
	shorttitle = {{NASimEmu}},
	urldate = {2023-11-20},
	publisher = {arXiv},
	author = {Janisch, Jaromír and Pevný, Tomáš and Lisý, Viliam},
	month = aug,
	year = {2023},
	note = {arXiv:2305.17246 [cs]},
}

@article{sarraute_pomdps_2012,
	title = {{POMDPs} {Make} {Better} {Hackers}: {Accounting} for {Uncertainty} in {Penetration} {Testing}},
	volume = {26},
	copyright = {Copyright (c) 2021 Proceedings of the AAAI Conference on Artificial Intelligence},
	issn = {2374-3468},
	shorttitle = {{POMDPs} {Make} {Better} {Hackers}},
	language = {en},
	number = {1},
	urldate = {2023-11-21},
	journal = {Proceedings of the AAAI Conference on Artificial Intelligence},
	author = {Sarraute, Carlos and Buffet, Olivier and Hoffmann, Jörg},
	year = {2012},
	note = {Number: 1},
	pages = {1816--1824},
}

@article{li_eppta_2023,
	title = {{EPPTA}: Efficient partially observable reinforcement learning agent for penetration testing applications},
	volume = {n/a},
	rights = {© 2023 The Authors. Engineering Reports published by John Wiley \& Sons Ltd.},
	issn = {2577-8196},
	shorttitle = {{EPPTA}},
	pages = {e12818},
	issue = {n/a},
	journaltitle = {Engineering Reports},
	author = {Li, Zegang and Zhang, Qian and Yang, Guangwen},
	urldate = {2024-07-03},
	langid = {english},
	note = {\_eprint: https://onlinelibrary.wiley.com/doi/pdf/10.1002/eng2.12818},
    year = {2023}
}

@misc{sarraute_penetration_2013,
	title = {Penetration {Testing} == {POMDP} {Solving}?},
	language = {en},
	urldate = {2023-12-12},
	publisher = {arXiv},
	author = {Sarraute, Carlos and Buffet, Olivier and Hoffmann, Joerg},
	month = jun,
	year = {2013},
	note = {arXiv:1306.4714 [cs]},
}

@misc{msft:cyberbattlesim,
	Author = {Microsoft Defender Research Team.},
	Note = {Created by Christian Seifert, Michael Betser, William Blum, James Bono, Kate Farris, Emily Goren, Justin Grana, Kristian Holsheimer, Brandon Marken, Joshua Neil, Nicole Nichols, Jugal Parikh, Haoran Wei.},
	Publisher = {GitHub},
	Howpublished = {\url{https://github.com/microsoft/cyberbattlesim}},
	Title = {CyberBattleSim},
	Year = {2021}
}

@misc{schwartz2019nasim,
	title={NASim: Network Attack Simulator},
	author={Schwartz, Jonathon and Kurniawatti, Hanna},
	year={2019},
	howpublished={\url{https://networkattacksimulator.readthedocs.io/}},
}

@book{sutton_rl_2018,
	author = {Sutton, Richard S. and Barto, Andrew G.},
	title = {Reinforcement Learning: An Introduction},
	year = {2018},
	isbn = {0262039249},
	publisher = {A Bradford Book},
	address = {Cambridge, MA, USA}
}

@article{dulac2021challenges,
	title={Challenges of real-world reinforcement learning: definitions, benchmarks and analysis},
	author={Dulac-Arnold, Gabriel and Levine, Nir and Mankowitz, Daniel J and Li, Jerry and Paduraru, Cosmin and Gowal, Sven and Hester, Todd},
	journal={Machine Learning},
	volume={110},
	number={9},
	pages={2419--2468},
	year={2021},
	publisher={Springer}
}

@article{mnih2015human,
	title={Human-level control through deep reinforcement learning},
	author={Mnih, Volodymyr and Kavukcuoglu, Koray and Silver, David and Rusu, Andrei A and Veness, Joel and Bellemare, Marc G and Graves, Alex and Riedmiller, Martin and Fidjeland, Andreas K and Ostrovski, Georg and others},
	journal={Nature},
	volume={518},
	number={7540},
	pages={529--533},
	year={2015},
	publisher={Nature Publishing Group}
}

@article{huang_closer_2022,
	title = {A {Closer} {Look} at {Invalid} {Action} {Masking} in {Policy} {Gradient} {Algorithms}},
	volume = {35},
	issn = {2334-0762},
	language = {en},
	urldate = {2023-12-19},
	journal = {The International FLAIRS Conference Proceedings},
	author = {Huang, Shengyi and Ontañón, Santiago},
	month = may,
	year = {2022},
	note = {arXiv:2006.14171 [cs, stat]},
}

@article{openai2019dota,
  title={Dota 2 with Large Scale Deep Reinforcement Learning},
  author={OpenAI and Christopher Berner and Greg Brockman and Brooke Chan and Vicki Cheung and Przemysław Dębiak and Christy Dennison and David Farhi and Quirin Fischer and Shariq Hashme and Chris Hesse and Rafal Józefowicz and Scott Gray and Catherine Olsson and Jakub Pachocki and Michael Petrov and Henrique Pondé de Oliveira Pinto and Jonathan Raiman and Tim Salimans and Jeremy Schlatter and Jonas Schneider and Szymon Sidor and Ilya Sutskever and Jie Tang and Filip Wolski and Susan Zhang},
  year={2019},
  eprint={1912.06680},
  archivePrefix={arXiv},
  url={https://arxiv.org/abs/1912.06680}
}

@article{patterson2024empirical,
  title={Empirical design in reinforcement learning},
  author={Patterson, Andrew and Neumann, Samuel and White, Martha and White, Adam},
  journal={Journal of Machine Learning Research},
  volume={25},
  number={318},
  pages={1--63},
  year={2024}
}

@article{huang2022cleanrl,
  author  = {Shengyi Huang and Rousslan Fernand Julien Dossa and Chang Ye and Jeff Braga and Dipam Chakraborty and Kinal Mehta and João G.M. Araújo},
  title   = {CleanRL: High-quality Single-file Implementations of Deep Reinforcement Learning Algorithms},
  journal = {Journal of Machine Learning Research},
  year    = {2022},
  volume  = {23},
  number  = {274},
  pages   = {1--18},
}

@inproceedings{cobbe_leveraging_2020,
author = {Cobbe, Karl and Hesse, Christopher and Hilton, Jacob and Schulman, John},
title = {Leveraging procedural generation to benchmark reinforcement learning},
year = {2020},
publisher = {JMLR.org},
booktitle = {Proceedings of the 37th International Conference on Machine Learning},
articleno = {191},
numpages = {9},
series = {ICML'20}
}

@inproceedings{cobbe2019quantifying,
  title={Quantifying generalization in reinforcement learning},
  author={Cobbe, Karl and Klimov, Oleg and Hesse, Chris and Kim, Taehoon and Schulman, John},
  booktitle={International conference on machine learning},
  pages={1282--1289},
  year={2019},
  organization={PMLR}
}

@incollection{puterman_1990_mdp,
title = {Chapter 8 Markov decision processes},
series = {Handbooks in Operations Research and Management Science},
publisher = {Elsevier},
volume = {2},
pages = {331-434},
year = {1990},
booktitle = {Stochastic Models},
issn = {0927-0507},
author = {Martin L. Puterman}
}

@techreport{nist800115,
  title={Technical Guide to Information Security Testing and Assessment},
  author={{NIST}},
  institution={NIST},
  number={SP 800-115},
  year={2008},
  month={September},
  url={https://doi.org/10.6028/NIST.SP.800-115}
}

@inproceedings{terranova2024leveraging,
  title={Leveraging Deep Reinforcement Learning for Cyber-Attack Paths Prediction: Formulation, Generalization, and Evaluation},
  author={Terranova, Franco and Lahmadi, Abdelkader and Chrisment, Isabelle},
  booktitle={Proceedings of the 27th International Symposium on Research in Attacks, Intrusions and Defenses},
  pages={1--16},
  year={2024}
}

@inproceedings{terranova2025scalable,
  title={Scalable and Generalizable RL Agents for Attack Path Discovery via Continuous Invariant Spaces},
  author={Terranova, Franco and Lahmadi, Abdelkader and Chrisment, Isabelle},
  booktitle={2025 28th International Symposium on Research in Attacks, Intrusions and Defenses (RAID)},
  pages={18},
  year={2025}
}

@software{jax2018github,
  author = {James Bradbury and Roy Frostig and Peter Hawkins and Matthew James Johnson and Chris Leary and Dougal Maclaurin and George Necula and Adam Paszke and Jake Vander{P}las and Skye Wanderman-{M}ilne and Qiao Zhang},
  title = {{JAX}: composable transformations of {P}ython+{N}um{P}y programs},
  url = {http://github.com/jax-ml/jax},
  version = {0.3.13},
  year = {2018},
}

@inproceedings{matthews2024craftax,
    author={Michael Matthews and Michael Beukman and Benjamin Ellis and Mikayel Samvelyan and Matthew Jackson and Samuel Coward and Jakob Foerster},
    title = {Craftax: A Lightning-Fast Benchmark for Open-Ended Reinforcement Learning},
    booktitle = {International Conference on Machine Learning ({ICML})},
    year = {2024}
}

@article{lu2022discovered,
    title={Discovered policy optimisation},
    author={Lu, Chris and Kuba, Jakub and Letcher, Alistair and Metz, Luke and Schroeder de Witt, Christian and Foerster, Jakob},
    journal={Advances in Neural Information Processing Systems},
    volume={35},
    pages={16455--16468},
    year={2022}
}

@misc{simon2025learningrobustpenetrationtestingpolicies,
      title={Learning Robust Penetration-Testing Policies under Partial Observability: A systematic evaluation}, 
      author={Raphael Simon and Pieter Libin and Wim Mees},
      year={2025},
      eprint={2509.20008},
      archivePrefix={arXiv},
      primaryClass={cs.LG},
      url={https://arxiv.org/abs/2509.20008}, 
}

@misc{bonnet2024jumanji,
    title={Jumanji: a Diverse Suite of Scalable Reinforcement Learning Environments in JAX},
    author={Clément Bonnet and Daniel Luo and Donal Byrne and Shikha Surana and Sasha Abramowitz and Paul Duckworth and Vincent Coyette and Laurence I. Midgley and Elshadai Tegegn and Tristan Kalloniatis and Omayma Mahjoub and Matthew Macfarlane and Andries P. Smit and Nathan Grinsztajn and Raphael Boige and Cemlyn N. Waters and Mohamed A. Mimouni and Ulrich A. Mbou Sob and Ruan de Kock and Siddarth Singh and Daniel Furelos-Blanco and Victor Le and Arnu Pretorius and Alexandre Laterre},
    year={2024},
    eprint={2306.09884},
    url={https://arxiv.org/abs/2306.09884},
    archivePrefix={arXiv},
    primaryClass={cs.LG}
}

@software{gymnax2022github,
  author = {Robert Tjarko Lange},
  title = {{gymnax}: A {JAX}-based Reinforcement Learning Environment Library},
  url = {http://github.com/RobertTLange/gymnax},
  version = {0.0.4},
  year = {2022},
}

@article{emerson2024cyborg++,
  title={Cyborg++: An enhanced gym for the development of autonomous cyber agents},
  author={Emerson, Harry and Bates, Liz and Hicks, Chris and Mavroudis, Vasilios},
  journal={arXiv preprint arXiv:2410.16324},
  year={2024}
}

@article{nguyen2025pengym,
title = {PenGym: Realistic training environment for reinforcement learning pentesting agents},
journal = {Computers \& Security},
volume = {148},
pages = {104140},
year = {2025},
issn = {0167-4048},
doi = {https://doi.org/10.1016/j.cose.2024.104140},
url = {https://www.sciencedirect.com/science/article/pii/S0167404824004450},
author = {Huynh Phuong Thanh Nguyen and Kento Hasegawa and Kazuhide Fukushima and Razvan Beuran},
}

@article{vinyals2019grandmaster,
  title={Grandmaster level in StarCraft II using multi-agent reinforcement learning},
  author={Vinyals, Oriol and Babuschkin, Igor and Czarnecki, Wojciech M and Mathieu, Micha{\"e}l and Dudzik, Andrew and Chung, Junyoung and Choi, David H and Powell, Richard and Ewalds, Timo and Georgiev, Petko and others},
  journal={nature},
  volume={575},
  number={7782},
  pages={350--354},
  year={2019},
  publisher={Nature Publishing Group}
}

@article{liu2025autonomous,
  title={Autonomous Penetration Testing using Reinforcement Learning: A Review and Perspectives},
  author={Liu, Jingju and Zhang, Yue and Zhou, Shicheng and Yang, Jiahai and Lu, Yuliang and Zhong, Xiaofeng},
  journal={Expert Systems with Applications},
  pages={130219},
  year={2025},
  publisher={Elsevier}
}

@article{hallak2015contextual,
  title={Contextual markov decision processes},
  author={Hallak, Assaf and Di Castro, Dotan and Mannor, Shie},
  journal={arXiv preprint arXiv:1502.02259},
  year={2015}
}

@inproceedings{jiang2021plr,
  title={Prioritized level replay},
  author={Jiang, Minqi and Grefenstette, Edward and Rockt{\"a}schel, Tim},
  booktitle={International Conference on Machine Learning},
  pages={4940--4950},
  year={2021},
  organization={PMLR}
}

@article{coward2024JaxUED,
  title={JaxUED: A simple and useable UED library in Jax},
  author={Samuel Coward and Michael Beukman and Jakob Foerster},
  journal={arXiv preprint},
  year={2024},
}

@INPROCEEDINGS{tobin_2017_DR,
  author={Tobin, Josh and Fong, Rachel and Ray, Alex and Schneider, Jonas and Zaremba, Wojciech and Abbeel, Pieter},
  booktitle={2017 IEEE/RSJ International Conference on Intelligent Robots and Systems (IROS)}, 
  title={Domain randomization for transferring deep neural networks from simulation to the real world}, 
  year={2017},
  volume={},
  number={},
  pages={23-30},
  doi={10.1109/IROS.2017.8202133}
}

@inproceedings{jiang_replay-guided_2021,
 author = {Jiang, Minqi and Dennis, Michael and Parker-Holder, Jack and Foerster, Jakob and Grefenstette, Edward and Rockt\"{a}schel, Tim},
 booktitle = {Advances in Neural Information Processing Systems},
 editor = {M. Ranzato and A. Beygelzimer and Y. Dauphin and P.S. Liang and J. Wortman Vaughan},
 pages = {1884--1897},
 publisher = {Curran Associates, Inc.},
 title = {Replay-Guided Adversarial Environment Design},
 url = {https://proceedings.neurips.cc/paper_files/paper/2021/file/0e915db6326b6fb6a3c56546980a8c93-Paper.pdf},
 volume = {34},
 year = {2021}
}

@InProceedings{parker-holder_2023_accel,
  title = 	 {Evolving Curricula with Regret-Based Environment Design},
  author =       {Parker-Holder, Jack and Jiang, Minqi and Dennis, Michael and Samvelyan, Mikayel and Foerster, Jakob and Grefenstette, Edward and Rockt{\"a}schel, Tim},
  booktitle = 	 {Proceedings of the 39th International Conference on Machine Learning},
  pages = 	 {17473--17498},
  year = 	 {2022},
  editor = 	 {Chaudhuri, Kamalika and Jegelka, Stefanie and Song, Le and Szepesvari, Csaba and Niu, Gang and Sabato, Sivan},
  volume = 	 {162},
  series = 	 {Proceedings of Machine Learning Research},
  month = 	 {17--23 Jul},
  publisher =    {PMLR},
  url = 	 {https://proceedings.mlr.press/v162/parker-holder22a.html}
}

@article{kirk_survey_2023,
	title = {A {Survey} of {Zero}-shot {Generalisation} in {Deep} {Reinforcement} {Learning}},
	volume = {76},
	issn = {1076-9757},
	url = {http://jair.org/index.php/jair/article/view/14174},
	doi = {10.1613/jair.1.14174},
	language = {en},
	urldate = {2025-10-30},
	journal = {jair},
	author = {Kirk, Robert and Zhang, Amy and Grefenstette, Edward and Rocktäschel, Tim},
	month = jan,
	year = {2023},
	pages = {201--264},
}

@article{adkins2024method,
  title={A method for evaluating hyperparameter sensitivity in reinforcement learning},
  author={Adkins, Jacob and Bowling, Michael and White, Adam},
  journal={Advances in Neural Information Processing Systems},
  volume={37},
  pages={124820--124842},
  year={2024}
}

@inproceedings{rutherford2024jaxmarl,
 author = {Rutherford, Alexander and Ellis, Benjamin and Gallici, Matteo and Cook, Jonathan and Lupu, Andrei and Ingvarsson, Gar\dh ar and Willi, Timon and Hammond, Ravi and Khan, Akbir and de Witt, Christian Schroeder and Souly, Alexandra and Bandyopadhyay, Saptarashmi and Samvelyan, Mikayel and Jiang, Minqi and Lange, Robert and Whiteson, Shimon and Lacerda, Bruno and Hawes, Nick and Rockt\"{a}schel, Tim and Lu, Chris and Foerster, Jakob},
 booktitle = {Advances in Neural Information Processing Systems},
 doi = {10.52202/079017-1612},
 editor = {A. Globerson and L. Mackey and D. Belgrave and A. Fan and U. Paquet and J. Tomczak and C. Zhang},
 pages = {50925--50951},
 publisher = {Curran Associates, Inc.},
 title = {JaxMARL: Multi-Agent RL Environments and Algorithms in JAX},
 url = {https://proceedings.neurips.cc/paper_files/paper/2024/file/5aee125f052c90e326dcf6f380df94f6-Paper-Datasets_and_Benchmarks_Track.pdf},
 volume = {37},
 year = {2024}
}

@article{van_hasselt2016learning,
  title={Learning values across many orders of magnitude},
  author={Van Hasselt, Hado P and Guez, Arthur and Hessel, Matteo and Mnih, Volodymyr and Silver, David},
  journal={Advances in neural information processing systems},
  volume={29},
  year={2016}
}

@article{gallici2025simplifying,
    title={Simplifying Deep Temporal Difference Learning},
    author={Matteo Gallici and Mattie Fellows and Benjamin Ellis
     and Bartomeu Pou and Ivan Masmitja and Jakob Nicolaus Foerster
      and Mario Martin},
    year={2025}, 
    eprint={2407.04811},
    journal={The International Conference on Learning Representations (ICLR)},
    primaryClass={cs.LG},
    url={https://arxiv.org/abs/2407.04811},
}

@article{zhou2024towards,
  title={Towards generalizable autonomous penetration testing via domain randomization and meta-reinforcement learning},
  author={Zhou, Shicheng and Liu, Jingju and Lu, Yuliang and Yang, Jiahai and Zhang, Yue and Chen, Jie},
  journal={ArXiv abs/2412.(04078)},
  year={2024}
}
